%% file: main.tex
\PassOptionsToPackage{table}{xcolor}
\documentclass[]{selfevolagent}

\usepackage{microtype}
\microtypesetup{expansion=false}
\usepackage{graphicx}
\usepackage{subcaption}
\usepackage{booktabs}
\usepackage{amsmath}
\usepackage{amsfonts}
\usepackage{amssymb}
\usepackage{nicefrac}
\usepackage{multirow}
\usepackage{url}
\usepackage{xspace}
\usepackage{tikz}
\usetikzlibrary{positioning, arrows.meta, shapes.geometric, fit, backgrounds, calc, patterns, tikzmark}
\usepackage{pgfplots}
\pgfplotsset{compat=1.18}
\usepgfplotslibrary{fillbetween}
\usepackage{algorithm}
\usepackage{algorithmic}
\usepackage{float}
\usepackage{fontawesome5}

\newcommand{\system}[0]{ProAct\xspace}
\newcommand{\benchmark}[0]{ProActEval\xspace}
\newcommand{\placeholderfigure}[1]{%
  \fbox{%
    \begin{minipage}[c][1.55in][c]{0.94\linewidth}
      \centering\footnotesize Temporary placeholder for \texttt{\detokenize{#1}}
    \end{minipage}%
  }%
}
\newcommand{\safeincludegraphics}[2][]{%
  \IfFileExists{#2}{\includegraphics[#1]{#2}}{\placeholderfigure{#2}}%
}
\newcommand{\corrauth}{\dagger}
\newcommand{\equalcontrib}{\ast}
\renewcommand\authorformat[2][]{{\normalsize\sffamily\bfseries #2\ifstrempty{#1}{}{$^{#1}$}}}

\title{Anticipate and Learn: Unleashing Idle-Time Compute in Proactive Agents}
\titlelogo{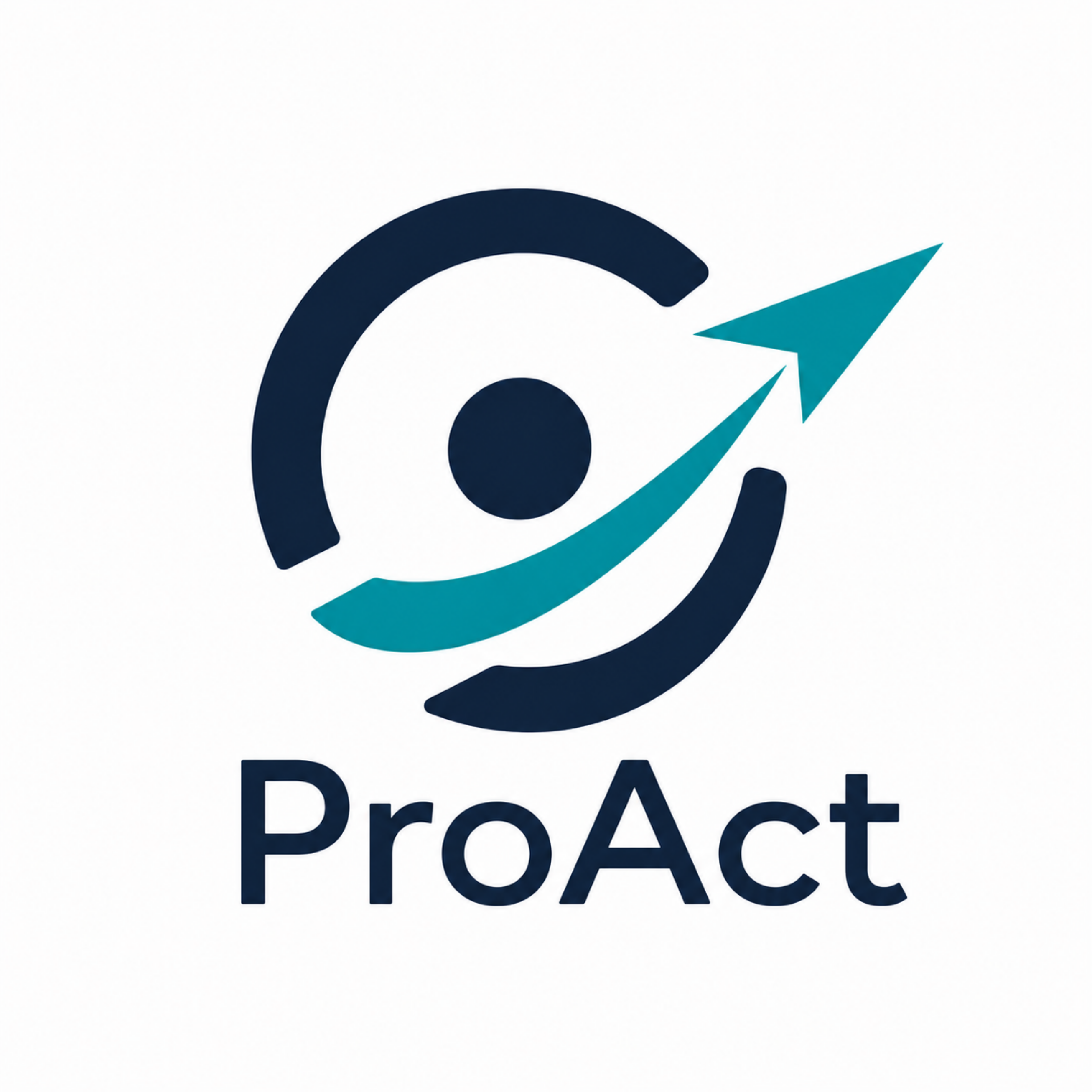}

\author[\equalcontrib,1]{Haoyi Hu}
\author[\equalcontrib,1]{~Qirong Lyu}
\author[1]{~Xianghan Kong}
\author[1,\corrauth]{~Weiwen Liu}
\author[1]{~Jianghao Lin}
\author[2]{~Zixuan Guo}
\author[2]{~Yan Xu}
\author[2]{~Yasheng Wang}
\author[1]{~Weinan Zhang}
\author[1]{~Yong Yu}

\affiliation[1]{Shanghai Jiao Tong University}
\affiliation[2]{Tencent}

\contribution[\equalcontrib]{~Equal contribution}
\contribution[\corrauth]{~Corresponding author}

\metadata[\raisebox{-0.18ex}{\scalebox{0.89}{\faEnvelope}}~Contact]{Weiwen Liu (\email{wwliu@sjtu.edu.cn})}
\metadata[{\faGithub}~Code]{\href{https://github.com/AgentACE-AI/ProAct}{\nolinkurl{https://github.com/AgentACE-AI/ProAct}}}

\abstract{
While AI agents demonstrate remarkable capabilities in reasoning and tool use, they remain fundamentally reactive: they compute responses only after explicit user prompts. This paradigm ignores a critical opportunity: the \textit{idle time} between interactions is largely wasted, leaving agents unable to prepare for future user needs. To bridge this gap, we introduce \system, a proactive agent architecture that leverages idle-time compute to anticipate and fulfill likely upcoming user needs. By analyzing evolving dialogue history together with persistent memory, \system predicts upcoming needs and iteratively acquires information, allowing the agent to resolve knowledge gaps and prepare evidence before the user initiates a query.
To rigorously evaluate proactive capabilities, we also introduce \benchmark, a comprehensive benchmark comprising 200 scenarios across 40 domains, featuring predictable need chains and diverse user cognitive profiles. Empirical results demonstrate significant advantages over reactive baselines. \system accelerates task completion by reducing required turns by 14.8\%, decreases user effort by 11.7\%, and cuts hallucination rates by 28.1\% on \benchmark. Furthermore, MemBench evaluations confirm that \system achieves state-of-the-art reflective accuracy, underscoring its sustained and robust performance.
}

\begin{document}

\maketitle

\input{sections/introduction}
\input{sections/related_work}

\input{sections/method_modified}
\input{sections/proactivebench}
\input{sections/experiments_modified}
\input{sections/conclusion}

\clearpage
\bibliographystyle{plainnat}
\bibliography{references}

\newpage
\beginappendix
\let\appendixsection\section
\renewcommand{\section}{\FloatBarrier\appendixsection}
\input{sections/appendix}
\FloatBarrier

\end{document}

%% file: sections/introduction.tex
% ====================================================================
% Section 1: Introduction
% ====================================================================

\section{Introduction}
\label{sec:intro}

Despite rapid advancements in conversational fluency, complex reasoning, and tool execution \citep{wang2024survey,wang2025toward,liu2024toolace}, today's deployed AI agents remain largely \textit{reactive and static} \citep{lu2024proactive}. They operate on a request-response basis, with processing initiated only after an explicit request is issued \citep{hu2024designing,lu2024proactive}. 
Consequently, once a task is completed, the agent returns to a dormant state. 
This design underutilizes potentially valuable idle time that could otherwise be used to refine the agent’s understanding of the user, anticipate probable future needs, and proactively prepare useful support for upcoming interactions \citep{wang2024survey}.

This limitation contrasts with the psychological concept of proactive coping \citep{greenglass1999proactive, drummond2016proactive}, a future-oriented strategy in which individuals anticipate upcoming demands, accumulate resources, and prepare for prospective goals before those demands fully materialize. Drawing on this distinction, we argue that \textbf{AI agents should view the idle time between user turns not as empty delay, but as an opportunity to anticipate, learn, and prepare for likely future demands}.

\begin{figure}[!t]
  \centering
  \safeincludegraphics[width=\linewidth]{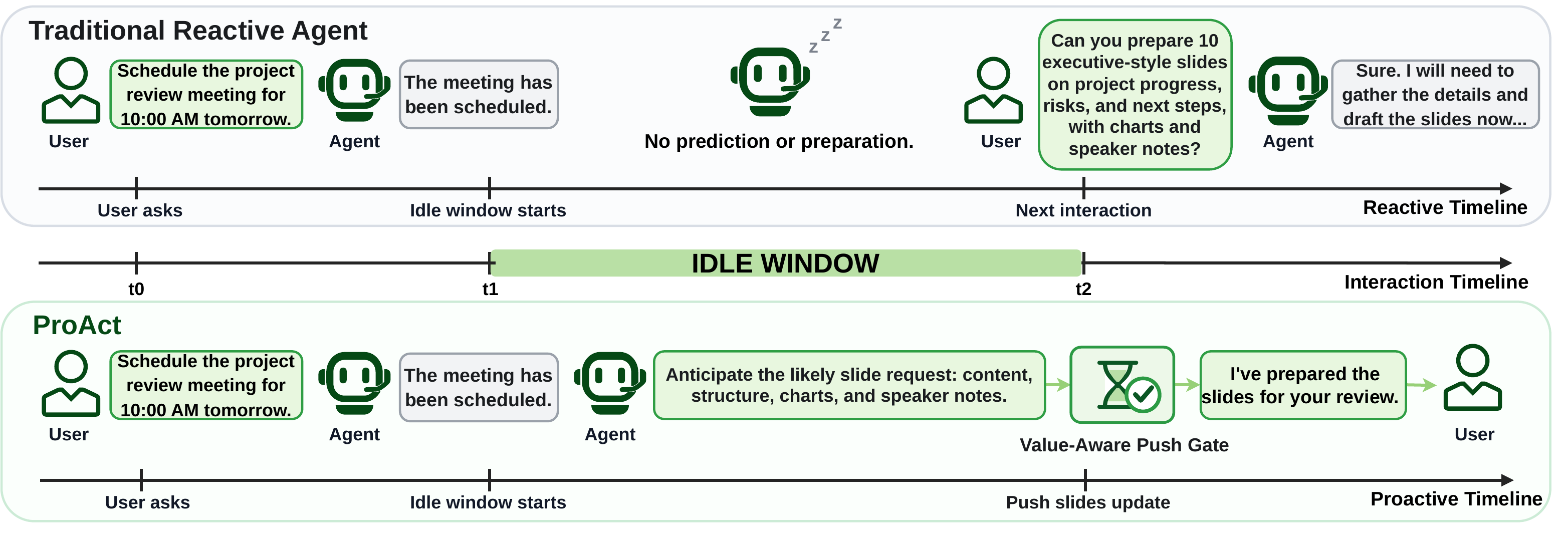}
  \vspace{-2em}
  \caption{Current assistants wait for explicit requests and leave idle-time compute unused. \system{} instead uses dialogue history and persistent memory to predict likely future needs, explores high-value candidates during idle windows, and feeds the resulting knowledge back into later interactions.}
  \label{fig:teaser}
  \vspace{-1.5em}
\end{figure}

This motivates a new human--AI collaboration paradigm: during the idle time between tasks, the agent continuously evolves rather than remaining static. Instead of concentrating all computation at the moment of interaction, the agent shifts substantial work into off-peak periods. From accumulated interaction history, the agent infers personalized preference patterns and future interests before they are explicitly requested. 
Figure~\ref{fig:teaser} illustrates this idea with a project-review scenario: after scheduling a meeting, a proactive agent can infer that review materials may soon be needed, prepare supporting content during the idle window, and deliver it only when a value-aware gate judges the intervention useful.
The core challenge, then, is how to transform idle time into useful proactive work without overwhelming the user with irrelevant, premature, or weakly grounded suggestions \citep{lu2024proactive, lin2025sleep}.

% We present \system, a unified architecture designed to address this challenge by turning idle time into a structured anticipating and learning procedure.
% \system is built around two tightly coupled modules.
% \textbf{Future-State Prediction} continuously anticipates the user's underlying future demands. Rather than relying solely on the most recent utterance, this module combines persistent user personas with the short-term dynamics of the active dialogue state. \textbf{Idle-Time Acquisition} then evaluates these predicted needs by expected user relevance, knowledge gap, incremental value, and timeliness, allocating background computation only to high-value candidates.
% For accepted candidates, the system retrieves and verifies supporting evidence, generates compact knowledge artifacts, and stores them in memory so they can be proactively delivered, woven into later responses, or silently reused when the user’s need materializes.

We present \system, a unified architecture that turns idle time into a structured cycle of anticipation and learning.  \system is driven by two tightly coupled modules. \textbf{Future-State Prediction} continuously forecasts the user's latent future demands. Rather than relying solely on the most recent utterance, this module integrates the dialogue history with persistent memory that captures user profiles, prior summaries, stored facts, and unresolved memory gaps to project likely upcoming intents. \textbf{Idle-Time Acquisition} subsequently evaluates these predicted needs based on expected user relevance, existing knowledge gaps, incremental value, and timeliness, judiciously allocating background computation only to high-value candidates. For these accepted candidates, the system retrieves and verifies supporting evidence, generates compact knowledge artifacts, and commits them to memory. Consequently, these insights can be proactively delivered, woven into subsequent responses, or silently retrieved the moment the user’s anticipated need materializes.

% \textbf{Future-State Prediction} forecasts likely upcoming user needs from historical interaction patterns, persistent user context, recent dialogue state, and unresolved knowledge gaps.

To evaluate idle-time compute for proactive agents, we introduce ProActEval, a 200-scenario, 40-domain evaluation framework with predictable need chains and diverse user cognitive profiles.
On ProActEval, \system accelerates task completion by reducing required turns by 14.8\%, decreases user effort by 11.7\%, and cuts hallucination rates by 28.1\% on \benchmark compared with a reactive baseline.
On MemBench, \system achieves 84.3\% reflective accuracy at 10k tokens and 86.3\% at 100k tokens.
This underscores the effectiveness of our idle-time compute for proactive agents, highlighting their potential to actively anticipate and learn to enhance user experiences.

Our core contributions are summarized as follows:
\begin{itemize}
\item We formulate a proactive human--AI collaboration paradigm and instantiate it in \system, an architecture that uses \textbf{Future-State Prediction} and \textbf{Idle-Time Acquisition} to turn idle intervals into grounded preparation for likely future needs.
\item We introduce \textbf{ProActEval}, a 200-scenario, 40-domain evaluation framework for benchmarking proactive agents with predictable need chains and diverse cognitive profiles.
\item We empirically validate \system, showing that it reduces interaction turns by 14.8\%, lowers user effort by 11.7\%, and mitigates hallucinations by 28.1\% on \benchmark, while achieving strong reflective accuracy on MemBench.
\end{itemize}

%% file: sections/related_work.tex
% ====================================================================
% Section 2: Related Work
% ====================================================================

\section{Related Work}
\label{sec:related}

\paragraph{Memory-augmented LLM agents.}
Several recent systems extend LLM agents with persistent memory.
Generative Agents \citep{park2023generative} maintain a memory stream with reflection and importance scoring but lack structured deduplication or lifecycle management.
MemGPT \citep{packer2023memgpt} introduces a virtual memory hierarchy inspired by operating systems, enabling paging between fast and archival memory; however, it does not model user profiles or support proactive behavior.
MemoryBank \citep{zhong2024memorybank} implements hierarchical daily summaries with an Ebbinghaus forgetting mechanism but operates strictly on demand.
SCMemory \citep{wang2023enhancing} proposes self-controlled memory selection but remains reactive.
GAM \citep{yan2025general} further reframes memory as just-in-time context construction but remains primarily request-driven and lacks proactive anticipation.
In contrast, \system{} unifies vector, relational, and document storage with an active knowledge lifecycle, incrementally updates user profiles and interaction-grounded facts, and couples memory directly to proactive behavior.

\paragraph{Proactive and anticipatory agents.}
Proactive computing has a long history in mobile and ubiquitous computing, but its integration with LLM-based agents is still early \citep{liao2023proactive}.
Recent work has explored proactive dialogue systems that predict user needs based on conversational context \citep{deng2023survey}, and self-reflective agents that trigger additional reasoning when uncertainty is high \citep{shinn2023reflexion, wang2023voyager}.
More recent agent systems such as OpenClaw and Hermes move toward always-on personal assistants, enabling scheduled checks, reminders, and automated task execution.
However, their proactive behavior is still largely initiated through user-specified schedules, routines, or explicit automation instructions, rather than through autonomous anticipation of unstated future needs.
These systems therefore remain limited in two ways: they either rely on the current conversational context to decide when to act, or depend on user-defined triggers after deployment.
In contrast, \system{} proactively infers future information needs without requiring users to predefine tasks or schedules.
Its proactive pipeline uses long-term user grounding, value-aware evaluation that balances information utility against interruption cost, and incremental research that reuses prior findings.

\paragraph{Inference-time compute.}
Another line of work improves LLM agents by allocating additional computation to planning, reflection, or iterative refinement at inference time. 
Self-reflective agents use feedback from past attempts to improve future actions, and recent test-time computation methods show that additional reasoning can improve performance on difficult tasks \citep{lin2025sleep, zhang2026lightweight, gupta2024metareflection, gao2025survey}.
However, these methods remain reactive: additional computation is triggered only after a user has issued a request, and is used to improve the response to that request rather than to anticipate and prepare for future user needs during idle periods.
\system{} instead treats background computation as a proactive mechanism, it predicts likely future needs, evaluates whether acting on them is worthwhile, and incrementally prepares grounded
assistance using long-term memory and prior findings.

%% file: sections/method_modified.tex
\section{Method}
\label{sec:method}

\subsection{Overview}
\label{sec:method:overview}

\begin{figure}[!t]
  \centering
  \safeincludegraphics[width=\linewidth]{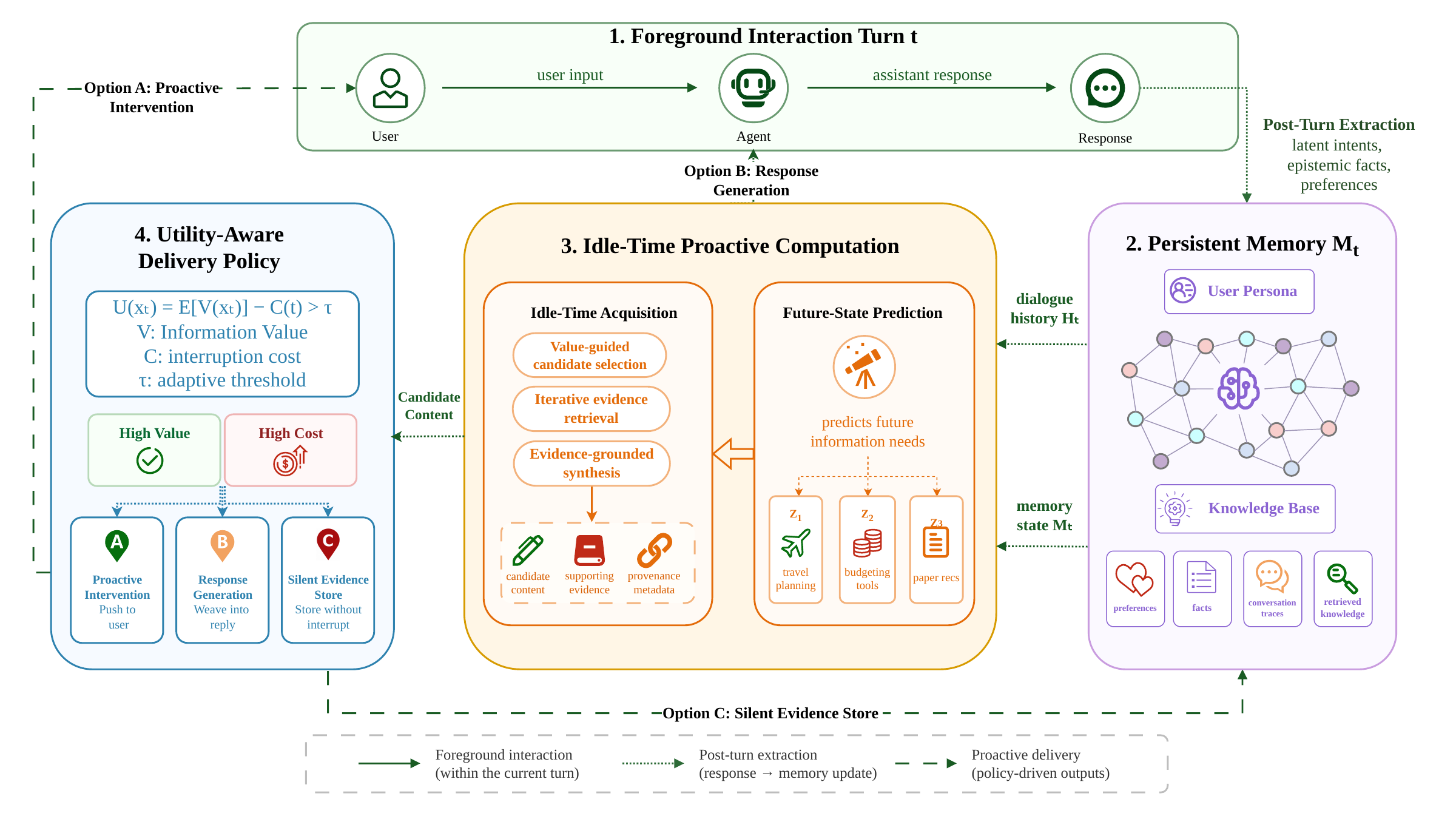}
  \caption{\system{} overview. After each foreground interaction, the agent updates persistent memory, predicts likely future needs during idle intervals, and acquires evidence for high-value candidates. A utility-aware delivery policy then handles the resulting artifacts for future use.}
  \label{fig:architecture}
\end{figure}

\system{} is designed for multi-turn settings in which the dialogue history and persistent memory state make some future information needs predictable.
Instead of waiting for an explicit request, the assistant uses this state to predict follow-up needs and prepare supporting evidence during idle intervals.
Figure~\ref{fig:architecture} summarizes this loop: foreground interactions update memory, which then conditions prediction, acquisition, and delivery decisions in the following idle interval.

The memory layer maintains user profiles, entity-level facts, conversation summaries, and acquired artifacts.
During an idle interval, \textbf{Future-State Prediction} generates a compact set of candidate future needs from the dialogue history and persistent memory state.
\textbf{Idle-Time Acquisition} scores the predicted needs, allocates idle-time budget to candidates worth additional computation, and performs evidence search or artifact generation when external support is needed.
A delivery policy then decides whether an artifact should be pushed immediately, queued for later use, or stored silently in memory.

\subsection{Proactive Agent Formulation}
\label{sec:method:formulation}

We formulate proactive agent interaction as a closed-loop decision problem.
After each foreground interaction, the agent updates its memory, predicts possible future needs, allocates idle-time computation to valuable candidates, and decides how the resulting preparation should be handled.
This formulation ties prediction, acquisition, and delivery to a single policy, rather than treating idle-time compute as unconstrained background search.

Let \(H_t = \{(u_1,a_1), \ldots, (u_t,a_t)\}\)
denote the dialogue history up to turn \(t\), where \(u_i\) and \(a_i\) are the user message and assistant response at turn \(i\).
Let \(M_t\) be the persistent memory state before idle-time computation.
After the current response, the system may receive an idle window \(\Delta_t\) with computation or retrieval budget \(B_t\).
In the meeting-schedule example in Figure~\ref{fig:teaser}, \(H_t\) contains the recent scheduling exchange, while \(M_t\) may contain remembered project context such as progress updates, risks, milestones, or prior artifacts.

The predictor generates a set of possible future needs:
\[
\mathcal{Z}_t = f_{\mathrm{pred}}(H_t, M_t).
\]
Each candidate \(z \in \mathcal{Z}_t\) is represented as
\[
z = (q_z, e_z, c_z, \rho_z),
\]
where \(q_z\) is the anticipated need, \(e_z\) is the grounding rationale from \(H_t\) or \(M_t\), \(c_z\) is the prediction confidence, and \(\rho_z\) is the retrieval plan used if the candidate is selected for acquisition.
For instance, a likely request for review materials can be grounded in the scheduled meeting and remembered project state, while its retrieval plan can point to relevant progress, risk, milestone, or metric evidence.

Given \((H_t,M_t,\Delta_t,B_t)\), the proactive policy \(\pi\) selects candidates, allocates budget, generates artifacts when useful, and assigns each prepared artifact a delivery decision \(d_z\).
The policy is optimized for future utility under interruption, budget, and factuality constraints:
\[
\max_{\pi}\; \mathbb{E}\Big[
U_{\mathrm{future}}(\pi; H_t,M_t)
- \lambda_i C_{\mathrm{interrupt}}(\pi)
- \lambda_b C_{\mathrm{budget}}(\pi)
- \lambda_h R_{\mathrm{hallucination}}(\pi)
\Big].
\]
Here, \(U_{\mathrm{future}}\) denotes the expected benefit of proactive preparation, such as reduced user effort, higher coverage, or faster completion.
\(C_{\mathrm{interrupt}}\), \(C_{\mathrm{budget}}\), and \(R_{\mathrm{hallucination}}\) denote interruption cost, computation cost, and hallucination risk, respectively.
Their weights \(\lambda_i\), \(\lambda_b\), and \(\lambda_h\) control the corresponding trade-offs.

Because downstream utility is not directly observable during idle intervals, \system{} uses a candidate-level value score for acquisition gating:
\[
S(z) = w_r r_z + w_g g_z + w_v v_z + w_{\tau} \tau_z,
\qquad
w_r + w_g + w_v + w_{\tau} = 1.
\]
Here, \(r_z\) measures user relevance, \(g_z\) measures the knowledge gap, \(v_z\) measures incremental value beyond existing memory, and \(\tau_z\) measures timeliness.
The weights \(w_r,w_g,w_v,w_{\tau}\) specify their relative importance.
This score is used by Idle-Time Acquisition to decide which predicted needs are worth preparing for.

\subsection{Future-State Prediction}
\label{sec:method:fsp}

Future-State Prediction instantiates \(f_{\mathrm{pred}}\) in Section~\ref{sec:method:formulation}.
Rather than expanding the search space broadly, it constructs a compact candidate set \(\mathcal{Z}_t\) whose members are traceable to the current dialogue, persistent memory, or identified memory gaps.
In the meeting-schedule example, this means predicting needs that naturally follow from the upcoming review, such as preparing progress summaries, risk updates, or supporting evidence.

\paragraph{Candidate generation.}
The predictor generates candidates from two sources.
First, local scenario prediction extrapolates near-term follow-up needs from the recent turns and immediate task reflected in \(H_t\).
Second, related expansion proposes adjacent topics grounded in \(M_t\), including user profiles, conversation summaries, stored artifacts, and unresolved goals.
The former captures needs directly implied by the current interaction, while the latter supports longer-range preparation based on stable user interests or ongoing projects.

\paragraph{Memory-gap augmentation.}
The predictor also receives signals from memory maintenance.
When the memory layer identifies stale, incomplete, weakly supported, or missing knowledge, these gaps are converted into candidate future needs and added to \(\mathcal{Z}_t\).
This allows memory maintenance to shape acquisition targets, instead of serving only as passive storage.

\paragraph{Filtering and prioritization.}
The raw candidate set is filtered by confidence and deduplicated against artifacts already stored in \(M_t\).
Candidates with confidence below \(\theta_{\mathrm{conf}}\) are removed.
The remaining candidates are grouped by topic similarity and prioritized, reducing near-duplicate exploration while preserving distinct future directions.
The output is the structured set \(\mathcal{Z}_t\) passed to Idle-Time Acquisition.

\subsection{Idle-Time Acquisition and Delivery}
\label{sec:method:ite}

Idle-Time Acquisition implements the acquisition and delivery components of the policy \(\pi\).
Given the predicted candidates \(\mathcal{Z}_t\), it applies the value gate from Section~\ref{sec:method:formulation}, checks memory coverage, acquires missing evidence when needed, and routes the resulting artifacts for later use.

\paragraph{Value evaluation.}
For each candidate \(z\), the module computes the value score \(S(z)\).
A candidate is acquired only if
\(S(z) \geq \theta_{\mathrm{val}}\).
Candidates below the threshold may be retained for later consideration, but they do not consume immediate evidence-search or artifact-generation budget.

\paragraph{Memory-aware acquisition.}
For accepted candidates, the module first checks whether the existing memory state \(M_t\) already contains sufficient evidence.
If memory coverage is high, the system reuses stored evidence and avoids redundant search.
If coverage is partial, it searches only for missing subtopics.
If coverage is low, it decomposes the candidate into sub-questions and performs iterative search, evidence extraction, and coverage checking.
This makes idle-time acquisition incremental rather than a full restart for every predicted need.

\paragraph{Artifact generation.}
Retrieved or remembered evidence is used to generate a compact knowledge artifact \(A_z\).
Each artifact contains the candidate need it supports, a preparation note, and provenance linking it to remembered or retrieved evidence.
This provenance allows proactively prepared content to be reused in later responses without weakening factual grounding.

\paragraph{Utility-aware delivery.}
After each artifact generation, the delivery policy selects a delivery mode \(d_z \in \{ \mathrm{push}, \mathrm{queue}, \mathrm{store} \}\).
An artifact is pushed only when its expected future utility justifies the interruption cost.
If it is useful but not urgent, it is queued for integration into a later response.
If it is potentially useful but not appropriate for immediate delivery, it is stored silently in memory.
This gate separates proactive assistance from background accumulation: prepared knowledge is acted on only when doing so is expected to help the user.

\paragraph{Memory update.}
After acquisition and delivery decisions, each artifact and its provenance are written back into memory, allowing later predictions and responses to reuse grounded preparation.
The resulting loop is
\[
(H_t, M_t)
\rightarrow
\mathcal{Z}_t
\rightarrow
S(z)
\rightarrow
A_z
\rightarrow
d_z
\rightarrow
M_{t+1}.
\]
Thus, memory serves as the shared state that couples prediction, acquisition, delivery, and future response generation.

%% file: sections/proactivebench.tex
% ====================================================================
% Section 4: ProActEval
% ====================================================================

\section{ProActEval}
\label{sec:proactivebench}

Evaluating proactive agents requires more than testing whether a system can answer the current question. Existing memory benchmarks~\citep{tan2025membench, zhang2024memsim, wu2024longmemeval, du2024perltqa, kim2024dialsim} primarily evaluate reactive recall or long-term question answering, while proactive benchmarks~\citep{lu2024proactive, de2026proactivebench} focus on task prediction from activity traces rather than memory-grounded anticipation in conversation.
A benchmark for this setting must specify which future needs are reasonably predictable, which facts ground those needs, and when proactive delivery should reduce later user effort.
We introduce ProActEval, an evaluation framework with 200 scenarios across 40 domains, fictional entities, scenario-specific fact sheets, and predictable need chains that measure whether agents can anticipate future conversational needs, reduce user effort, and maintain factual integrity.

\subsection{Benchmark Construction}
\label{sec:bench:construction}

Each ProActEval scenario is built around a self-contained fact sheet and an ordered set of user needs.
The fact sheet contains atomic facts with stable identifiers.
All scenario-specific entities, including people, organizations, addresses, dates, emails, and internal URLs, are fictional.
This controlled setup supports auditable factual evaluation: a response is correct only when it can be traced to the provided facts, and unsupported content is counted as hallucination.

The user needs define the interaction structure.
Each need has an importance label, one or more grounding fact identifiers, and a turn order.
Some needs also contain a \texttt{predictable\_after} field indicating that the need becomes reasonably anticipatable after earlier needs have been addressed.
Needs are organized into reveal groups to model local topic structure and topic shifts.
Together, these annotations form a user-needs graph.
The assistant cannot see the graph at runtime, but the simulator and evaluator use it to determine when proactive coverage should reduce future user effort.

We organize scenarios around five cognitive archetypes:
Foundational Memory, Translation and Gap Resolution, Trace and Dependency Reasoning, Handoff and Consistency Control, and Readiness and Follow-through.
These archetypes are not task labels for the model.
They are construction controls that ensure the benchmark covers different forms of anticipatory demand, from recalling stable facts to preparing for delayed follow-up actions.

\subsection{Data Synthesis Pipeline}
\label{sec:bench:synthesis}

Scenario generation proceeds in stages.
We begin with manually designed seed scenarios spanning personal life management, professional work, education, public services, finance, compliance, healthcare-adjacent support, and other specialized settings.
For each seed, we first generate the scenario-specific fact sheet and then generate the ordered user-need sequence conditioned on that fact sheet.
Separating fact generation from need generation makes it easier to audit grounding and predictability.

Generated scenarios are checked automatically for structural validity.
The checks enforce unique identifiers, legal fact references, acyclic predictability links, valid turn order, and reveal-group consistency.
For grouped scenarios, additional checks require enough cross-group predictability and enough auditable proactive targets so that the instance does not collapse into a purely reactive conversation.
After automatic validation, each scenario receives manual review for factual consistency, naturalness of need progression, plausibility of predictability links, and judge-friendliness.
Appendix~\ref{app:composition-stats} reports the benchmark composition statistics.

\subsection{Evaluation Protocol}
\label{sec:bench:protocol}

Each scenario-condition pair is evaluated with the same three-stage loop.
A user simulator traverses the ordered need sequence and emits a user message for the next unmet need.
If the assistant has already covered a future need proactively, the simulator skips that need, translating anticipation into reduced user effort.
The system under test then responds using only runtime-visible information: the user profile, the fact sheet loaded into memory or search, and the conversation history.
It never receives gold fields such as \texttt{user\_needs}, \texttt{key\_fact\_ids}, \texttt{predictable\_after}, or \texttt{reveal\_group}.
An LLM-based coverage judge finally marks which facts were correctly conveyed, which claims were distorted or unsupported, and which needs were addressed.

\subsection{Metrics}
\label{sec:bench:metrics}

We report seven metrics organized around efficiency, factual integrity, and coverage.
$T_{80}$ and $T_{100}$ measure the number of turns needed to reach 80\% and 100\% must-have coverage, respectively, while User Effort counts the turns in which the user explicitly asks a question.
Fact Accuracy measures the proportion of correctly conveyed facts among all facts delivered, and Hallucination Rate measures the proportion of unsupported claims.
Total Coverage and Must-Have Coverage measure the fraction of all needs and must-have needs satisfied by the end of the conversation, respectively.

%% file: sections/experiments_modified.tex
% ====================================================================
% Section 5: Experiments
% ====================================================================

\section{Experiments}
\label{sec:experiments}

% We evaluate \system{} on two benchmarks.
% MemBench measures whether the supporting memory layer can preserve reflective user information over long contexts.
% ProActEval measures whether idle-time proactive behavior reduces user effort, improves coverage, and preserves factual grounding.

We organize the evaluation around two main questions.
(1) We ask whether prediction-guided idle-time compute improves proactive assistance in multi-turn conversations.
We answer this question on \textbf{ProActEval}, where each scenario provides a fact sheet and an ordered sequence of user needs.
We compare three conditions: a reactive assistant, an undirected idle-time compute variant, and the full prediction-guided system.
This comparison evaluates the end-to-end benefit of proactive assistance while also ablating predictive direction: the gap between Undirected Idle and the full system tests whether background search alone is sufficient, or whether idle-time compute must be assigned to predicted needs. 
(2) We ask whether the memory backbone can reliably support such proactive behavior by preserving long-horizon user information.
We answer this question on \textbf{MemBench}, focusing on its reflective participation setting, which tests whether a system can infer user preferences and emotions from accumulated interaction history.
Finally, we extend the analysis by varying the idle-search budget on ProActEval to study the cost--efficiency trade-off of proactive computation.

\begin{table}[t]
\caption{Headline end-to-end results on the two evaluation suites.}
\label{tab:overall-main}
\centering
\small
\begin{tabular}{@{}l l l c c c@{}}
\toprule
\textbf{Suite} & \textbf{Evaluates} & \textbf{Primary endpoint} & \textbf{Baseline} & \textbf{\system{}} & \textbf{Change} \\
\midrule
MemBench & Reflective memory & 10k Acc. $\uparrow$ & 0.742 & \textbf{0.843} & +13.6\% \\
MemBench & Reflective memory & 100k Acc. $\uparrow$ & 0.833 & \textbf{0.863} & +3.6\% \\
ProActEval & Proactive assistance & $T_{100}$ $\downarrow$ & 8.110 & \textbf{6.910} & -14.8\% \\
ProActEval & Proactive assistance & User Effort $\downarrow$ & 9.140 & \textbf{8.075} & -11.7\% \\
\bottomrule
\end{tabular}
\end{table}

% \subsection{Benchmarks and Evaluation Targets}
% \paragraph{MemBench.}
% MemBench \citep{tan2025membench} evaluates LLM-based memory systems across factual and reflective tasks under participation and observation settings.
% We focus on the reflective participation setting because it tests whether a system can infer preferences and emotions from accumulated interaction history.
% This capability is important for proactive agents because prediction quality depends on persistent user context.
% \paragraph{ProActEval.}
% ProActEval evaluates proactive assistance in closed-world multi-turn conversations.
%Each scenario contains a fact sheet and an ordered sequence of user needs.
% The main outcome is not only whether the assistant is factually correct, but whether it can cover future-relevant needs before the user asks.
% We therefore report efficiency, coverage, anticipation, factual-integrity, and compute-cost metrics.

% \subsection{Main End-to-End Results}

% Table~\ref{tab:overall-main} summarizes the main end-to-end results on the two benchmarks using each benchmark's native primary endpoints.
% On MemBench, \system{} outperforms published structured-memory baselines at both context lengths.
% On ProActEval, the full proactive system improves interaction efficiency, coverage, and factual integrity; Fact Accuracy is reported in the detailed ProActEval table below.

\subsection{Main Proactivity Evaluation}

We evaluate proactive assistance on all 200 ProActEval scenarios by comparing a reactive baseline, an Idle-Time Acquisition variant without predictive direction, and the full prediction-guided \system{}.
\textsc{Reactive} is a non-proactive baseline that disables both \textsc{Future-State Prediction} and \textsc{Idle-Time Acquisition}.
\textsc{Undirected Idle} enables \textsc{Idle-Time Acquisition} but removes predictive direction, using unguided background intents.
\textsc{Directed Idle} is the full \system{} configuration, enabling both \textsc{Future-State Prediction} and \textsc{Idle-Time Acquisition}.
This design allows us to measure the end-to-end benefit of proactive assistance while isolating the contribution of predictive direction.

\paragraph{Overall Proactive Gains.}
Table~\ref{tab:proacteval-main} expands the ProActEval summary in Table~\ref{tab:overall-main} by reporting coverage, anticipation, factual integrity, and compute cost.
\textsc{Directed Idle} improves all non-cost metrics over both baselines, showing that prediction-guided idle-time computation improves not only turn efficiency but also proactive coverage and factual grounding.

\begin{table*}[t]
\caption{Detailed ProActEval results over 200 scenarios. \textsc{Reactive} disables both proactive modules; \textsc{Undirected Idle} enables idle-time acquisition without future-state prediction; \textsc{Directed Idle} is the full \system{} configuration. Delta columns report changes for \textsc{Directed Idle} relative to each baseline, except Anticipation Recall and Active Tokens, where deltas are absolute. Active Tokens are measured as additional active-token cost.}
\label{tab:proacteval-main}
\centering
\small
\resizebox{\textwidth}{!}{%
\begin{tabular}{@{}l c c c c c@{}}
\toprule
\textbf{Metric} & \textbf{Reactive} & \textbf{Undirected Idle} & \textbf{Directed Idle} & \textbf{$\Delta$ vs. Reactive} & \textbf{$\Delta$ vs. Undirected} \\
\midrule
\multicolumn{6}{l}{\textbf{Efficiency}} \\
\quad $T_{80}$ $\downarrow$              & 6.615 & 6.600 & \textbf{5.530} & $-16.4\%$ & $-16.2\%$ \\
\quad $T_{100}$ $\downarrow$             & 8.110 & 8.040 & \textbf{6.910} & $-14.8\%$ & $-14.1\%$ \\
\quad User Effort $\downarrow$           & 9.140 & 9.040 & \textbf{8.075} & $-11.7\%$ & $-10.7\%$ \\
\midrule
\multicolumn{6}{l}{\textbf{Coverage and Anticipation}} \\
\quad Total Coverage $\uparrow$          & 0.892 & 0.905 & \textbf{0.956} & $+7.2\%$ & $+5.6\%$ \\
\quad Must-Have Coverage $\uparrow$      & 0.938 & 0.950 & \textbf{0.977} & $+4.2\%$ & $+2.9\%$ \\
\quad Anticipation Recall $\uparrow$     & 0.000 & 0.000 & \textbf{0.428} & $+0.428$ & $+0.428$ \\
\midrule
\multicolumn{6}{l}{\textbf{Factual Integrity and Cost}} \\
\quad Fact Accuracy $\uparrow$           & 0.972 & 0.972 & \textbf{0.985} & $+1.3\%$ & $+1.3\%$ \\
\quad Hallucination Rate $\downarrow$    & 0.132 & 0.124 & \textbf{0.095} & $-28.1\%$ & $-23.1\%$ \\
\quad Active Tokens $\downarrow$         & \textbf{0}     & 69.8k & 111.8k & +111.8k & +42.0k \\
\bottomrule
\end{tabular}}
\end{table*}

\paragraph{Ablation Study.}
The comparison between \textsc{Undirected Idle} and \textsc{Directed Idle} isolates the value of predictive direction.
Although \textsc{Undirected Idle} performs Idle-Time Acquisition and spends 69.8k active tokens per scenario on average, it only slightly improves over \textsc{Reactive}, reducing $T_{100}$ by 0.9\% and User Effort by 1.1\%.
By contrast, adding predictive direction in \textsc{Directed Idle} reduces $T_{100}$ by 14.1\% and User Effort by 10.7\% relative to \textsc{Undirected Idle}, and achieves an Anticipation Recall of 0.428 where both other settings remain at 0.
These results suggest that the gains do not come from Idle-Time Acquisition alone, but from directing that acquisition toward predicted needs; undirected background compute adds cost but yields only limited proactive benefit.

\paragraph{Comparison with ProactiveAgent.}
We adapt the public ProactiveAgent decision protocol~\citep{lu2024proactive} as a GPT-4o prompting baseline. Since this baseline sees the user profile and full fact sheet but does not expose structured delivered fact IDs, we compare systems using judge-labeled Anticipation Recall rather than runtime fact-ID matching.

\begin{table*}[t]
\caption{Comparison with ProactiveAgent on 200 ProActEval scenarios. Judge-labeled Ant.R. is the scenario-level macro average; Anticipated Needs reports the corresponding micro count over all predictable needs.}
\label{tab:proactiveagent-comparison}
\centering
\small
\resizebox{\textwidth}{!}{%
\begin{tabular}{@{}l c c c c c@{}}
\toprule
\textbf{Method} & \textbf{$T_{80}$} $\downarrow$ & \textbf{$T_{100}$} $\downarrow$ & \textbf{User Effort} $\downarrow$ & \textbf{Judge-labeled Ant.R.} $\uparrow$ & \textbf{Anticipated Needs} $\uparrow$ \\
\midrule
\textsc{ProactiveAgent} & 5.600 & 7.145 & 8.425 & 0.020 & 32 / 1,572 \\
\system{} (\textsc{Directed Idle}) & \textbf{5.530} & \textbf{6.910} & \textbf{8.075} & \textbf{0.447} & \textbf{703 / 1,572} \\
\bottomrule
\end{tabular}}
\end{table*}

Table~\ref{tab:proactiveagent-comparison} shows that ProactiveAgent's proactive behavior remains largely turn-local: it anticipates only 32 of 1,572 predictable needs, yielding 0.020 judge-labeled Anticipation Recall.
By contrast, \system{} anticipates 703 of 1,572 predictable needs, reaching 0.447 judge-labeled Anticipation Recall while also modestly improving turn efficiency over ProactiveAgent. This suggests that proactive behavior is not sufficient by itself: it must cover predictable, benchmark-relevant needs to reduce later user effort.

\subsection{Memory Evaluation}
\label{sec:exp:membench}

We evaluate whether the memory backbone can support proactive assistance using the reflective participation setting of MemBench at 10k and 100k token scales with Qwen2.5-7B-Instruct.
The main metric is reflective accuracy, which measures whether the system can infer user preferences and emotions from accumulated interaction history.
We compare against FullMemory, RecentMemory, RetrievalMemory, GenerativeAgent~\citep{park2023generative}, MemoryBank~\citep{zhong2024memorybank}, MemGPT~\citep{packer2023memgpt}, and SCMemory~\citep{wang2023enhancing}.
\system{} achieves the strongest reflective accuracy at both context lengths, improving over the best prior baseline from 0.742 to 0.843 at 10k tokens and from 0.833 to 0.863 at 100k tokens.
Detailed baseline comparisons, scenario-level results, and memory-efficiency measurements are reported in Appendix~\ref{app:membench-detail}.

\subsection{Search Budget Analysis}
\label{sec:exp:budget-scaling}

% We further vary the maximum number of idle-time compute on a matched 50-scenario subset.
% For each budget $k \in \{4,8,12,16\}$, we compare Directed Idle ($k$) against Undirected Idle ($k$).
% This sweep asks whether more idle-time compute monotonically improves end-to-end interaction quality.
% Figure~\ref{fig:budget-scaling} focuses on the matched-budget efficiency and cost endpoints: completion turns, explicit user effort, and active-token usage.

\begin{figure}[!t]
  \centering
  \safeincludegraphics[width=\linewidth]{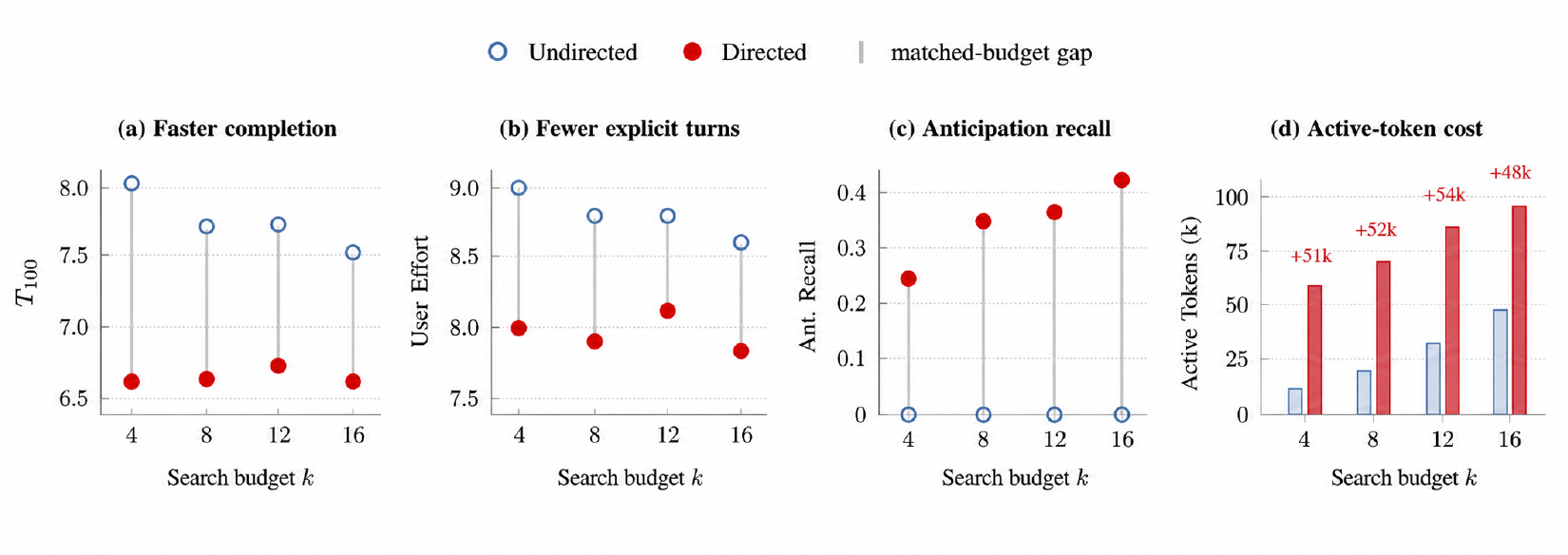}
  \caption{Search-budget analysis on a matched 50-scenario subset. Panels (a)--(c) compare \textsc{Directed Idle} and \textsc{Undirected Idle} under the same budget $k$, with gray segments denoting matched-budget gaps. Panel (d) reports active-token cost in thousands.}
  \label{fig:budget-scaling}
\end{figure}

\paragraph{Budget sweep setup.} We further study how the amount of idle-time acquisition affects proactive assistance.
On a matched 50-scenario subset, we vary the search budget $k \in \{4,8,12,16\}$ and compare \textsc{Directed Idle}$(k)$ with \textsc{Undirected Idle}$(k)$ under the same budget.
This controlled sweep tests whether increasing idle-time search monotonically improves interaction efficiency, and whether predictive direction remains beneficial when search volume is matched, and whether higher budgets continue to expand anticipation recall.
Figure~\ref{fig:budget-scaling} reports four endpoints: completion turns, explicit user effort, anticipation recall and active-token cost.

\paragraph{Cost--efficiency trade-off.}
Our experiment shows a clear cost--efficiency trade-off rather than a simple ``more search is better'' trend.
At every matched budget, \textsc{Directed Idle} achieves lower $T_{100}$ and lower User Effort than \textsc{Undirected Idle}, indicating that predictive direction improves the utility of idle-time acquisition beyond search volume alone.
Increasing $k$ continues to raise \textsc{Directed Idle}'s Anticipation Recall, from 0.253 at $k=4$ to 0.432 at $k=16$, but under the finite-scenario setting of ProActEval that extra recall does not guarantee monotonic reductions in User Effort.
Once the main predictable needs are covered, additional searches tend to chase lower-marginal or later needs, and each budget setting also induces a different closed-loop trajectory by changing memory contents, push timing, and subsequent dialogue context.
As a result, active-token cost rises steadily while $T_{100}$ and User Effort flatten or fluctuate.
Thus, the search budget should be treated as an operating point that balances efficiency gains against compute cost, rather than as a parameter to maximize. Higher Anticipation Recall does not translate linearly into fewer dialogue turns, because end-to-end efficiency can still be constrained by the last uncovered needs, simulator continuation policies, and strong full-context reactive baselines.

% \subsection{Discussion}
% \label{sec:exp:discussion}
% Overall, these experiments show that predictive direction is essential: undirected idle-time acquisition adds compute but yields limited proactive gains, whereas \system{} improves efficiency, coverage, and factual grounding. The ProactiveAgent comparison further suggests that frequent proactive attempts are insufficient unless they cover benchmark-labeled predictable needs. However, higher Anticipation Recall does not translate linearly into fewer dialogue turns, since end-to-end efficiency can still be limited by the last uncovered needs, simulator continuation policies, and strong full-context reactive baselines.

% The experiments support three conclusions.
% MemBench shows that \system{}'s memory layer is strong enough to support long-horizon personalization.
% ProActEval shows that proactive prediction combined with directed idle-time Acquisition improves end-to-end performance across efficiency, coverage, and factual-integrity metrics.
% The budget results also show that idle-time compute has a real cost.
% The best operating point should therefore be chosen by balancing efficiency gains, hallucination risk, and active-token budget.

%% file: sections/conclusion.tex
% ====================================================================
% Section 6: Conclusion
% ====================================================================

\section{Conclusion}
\label{sec:conclusion}

% We presented \system{}, a proactive agent architecture that uses future-state prediction to direct Idle-Time Acquisition.
% The central result is that Idle-Time compute becomes useful when it is assigned to likely future needs rather than spent on unguided background search.
% On ProActEval, \system{} reduces user effort by 11.7\%, accelerates task completion by reducing required turns by 14.8\%, increases total coverage by 7.2\%, and reduces hallucination rate by 28.1\% over a reactive baseline.
% Against undirected Idle-Time Acquistion, it still reduces user effort by 10.7\% and complete-coverage turns by 14.1\%.
% Budget-scaling results further show that predictive direction improves efficiency at matched budgets, while larger Idle-Time compute budgets introduce cost and diminishing returns.
% These findings suggest that proactive agents should be evaluated not only by answer quality, but also by how effectively they turn idle time into timely, grounded assistance.

We presented \system{}, a proactive agent architecture that uses persistent memory, Future-State Prediction, and Idle-Time Acquisition to convert idle intervals into grounded preparation for likely future needs. Across ProActEval and MemBench, \system{} improves proactive efficiency, coverage, factual integrity, and reflective memory accuracy. Our budget analysis further shows that larger Idle-Time Acquisition budgets raise active-token cost and yield diminishing returns, so proactive computation is an operating-point trade-off rather than something to maximize.

\paragraph{Limitations.}
These results come from a closed-world synthetic benchmark, so they should be read as controlled evidence for prediction-guided idle-time compute rather than a deployment guarantee in open-world personal assistants. The evaluation also depends on an LLM judge and on a value-aware delivery gate, and proactive preparation can still backfire when it competes with the reactive answer or pushes low-value content. Real deployments would need user controls, rate limits, and ongoing monitoring.

%% file: sections/appendix.tex
\section{Scenario Anatomy}
\label{app:scenario-anatomy}

Each ProActEval scenario separates runtime-visible information from judge-visible metadata.
The fact sheet provides scenario-specific ground truth, while the hidden user-needs graph records \texttt{key\_fact\_ids}, importance labels, reveal groups, and \texttt{predictable\_after} dependencies.
The system under test may use the user profile, fact sheet, and conversation history.
It does not receive gold user needs, grounding fact IDs, predictability links, or reveal-group annotations.
Those fields are used only by the user simulator and coverage judge.
This separation is necessary because proactive delivery should be measured against future needs without leaking those needs to the assistant.

\section{Scenario Data Example}
\label{app:scenario-data}

To make the scenario anatomy concrete, we reproduce excerpts from the \texttt{finance\_basic\_01} scenario in the \texttt{financial\_planning} domain.
The full scenario contains 28 facts and 12 user needs across 8 reveal groups.
All entities are fictional.

\paragraph{User profile.}
The simulated user is a 23-year-old entry-level analyst who just received their first paycheck and wants simple explanations about saving and investing.

\paragraph{Fact sheet (excerpt).}
Each fact is an atomic, verifiable statement with a unique identifier and category label.
\begin{small}
\begin{verbatim}
F06  [high_yield_savings_account]
     The Apex High-Yield Savings account from
     Apex Digital Bank offers an APY of 4.50%.

F07  [high_yield_savings_account]
     The Apex High-Yield Savings account has
     no monthly maintenance fees.

F14  [index_fund]
     The G500 fund has an expense ratio of
     0.04%, which is an annual fee.

F20  [retirement_account_401k]
     Innovatech Solutions provides a 100% match
     on 401k contributions up to 4% of pre-tax
     salary.

F26  [retirement_account_401k]
     Employees can enroll in the 401k plan at
     any time through the portal:
     https://my.horizonretirement.com/innovatech.
\end{verbatim}
\end{small}

\paragraph{User needs (excerpt).}
Each need specifies its importance level, grounding fact IDs, and predictability chain.
The \texttt{predictable\_after} field defines which earlier need makes this one anticipatable.

\begin{table}[!htbp]
\caption{Selected user needs from \texttt{finance\_basic\_01}. Needs marked with \texttt{predictable\_after} can in principle be anticipated after the referenced need is covered. The system under test never sees this metadata.}
\label{tab:scenario-needs}
\centering
\scriptsize
\resizebox{\linewidth}{!}{%
\begin{tabular}{@{}llllll@{}}
\toprule
\textbf{ID} & \textbf{Description} & \textbf{Level} & \textbf{Key facts} & \textbf{Predictable after} & \textbf{Reason} \\
\midrule
N1 & What is my company's 401k match? & must-have & F20 & -- & -- \\
N2 & What is the APY for the Apex HYSA? & must-have & F06 & -- & -- \\
N3 & What are the fees and min.\ deposit for the HYSA? & must-have & F07, F08 & N2 & After learning the rate, the user asks about costs to open. \\
N6 & How do I enroll in the 401k? & must-have & F26, F27 & N1 & Cross-group: learning the match triggers the enrollment question. \\
N9 & What is the vesting schedule? & nice-to-have & F21 & N1 & Cross-group: the match prompts questions about keeping it. \\
\bottomrule
\end{tabular}}
\end{table}

\paragraph{Reveal groups.}
The 12 needs are organized into 8 reveal groups (e.g., \texttt{G1: 401k\_match\_policy} $\to$ \{N1\}, \texttt{G4: 401k\_enrollment} $\to$ \{N6\}).
Groups with a \texttt{trigger\_after} dependency (e.g., G4 triggers after G1) model cross-topic anticipation targets: once the user learns about the 401k match (G1), enrollment logistics (G4) and vesting conditions (G6) become predictable.

\section{End-to-End Operational Example}
\label{app:operational-example}

This section expands the running project-review example from Figure~\ref{fig:teaser} into the concrete pipeline stages used by \system{}.
All thresholds and weights below correspond to the implementation defaults:
confidence filter $\theta_{\mathrm{conf}}=0.6$,
value gate $\theta_{\mathrm{val}}=60$ on a 0--100 scale (dividing by 100 gives the normalized score in Section~\ref{sec:method:formulation}),
scoring weights $w_r = w_g = w_v = w_\tau = 0.25$,
push-notification threshold 40,
and high-priority push threshold 70.

\paragraph{Runtime state.}
Suppose the user asks the assistant to schedule a project review meeting for 10:00 AM the next day.
After the scheduling response, the foreground turn is complete and the agent enters an idle window.
The runtime-visible state contains the recent request, a persistent memory summary for the project, progress notes from previous turns, a risk log, milestone records, and quantitative status metrics.
The system does not know the user's future request, but the scheduled review and project memory make several future needs predictable.

\paragraph{Candidate generation.}
Future-State Prediction generates candidates from two sources: local scenario prediction from the current dialogue (Section~\ref{sec:method:fsp}) and related expansion from persistent memory.
Memory maintenance may also add candidates for stale or incomplete project knowledge.
The implementation returns at most three predictor candidates with confidence at least $\theta_{\mathrm{conf}}=0.6$, while memory-gap candidates are added separately when the memory critic finds missing or weakly supported information.
Table~\ref{tab:operational-candidates} shows one possible candidate set for the idle window.
Each row corresponds to a predicted need $z = (q_z, e_z, c_z, \rho_z)$ from the formulation in Section~\ref{sec:method:formulation}.

\begin{table}[!htbp]
\caption{Example candidate future needs generated after the project-review scheduling turn. Columns map to the candidate structure in Section~\ref{sec:method:formulation}: Candidate need $= q_z$, Grounding rationale $= e_z$, Conf.\ $= c_z$, Retrieval plan $= \rho_z$.}
\label{tab:operational-candidates}
\centering
\scriptsize
\resizebox{\linewidth}{!}{%
\begin{tabular}{@{}p{0.22\linewidth}p{0.13\linewidth}p{0.33\linewidth}p{0.24\linewidth}c@{}}
\toprule
\textbf{Candidate need} ($q_z$) & \textbf{Source} & \textbf{Grounding rationale} ($e_z$) & \textbf{Retrieval plan} ($\rho_z$) & \textbf{Conf.} ($c_z$) \\
\midrule
Prepare executive-style review slides & \texttt{scenario} & A scheduled project review usually requires a concise progress, risk, and next-step briefing for stakeholders. & Retrieve progress notes, risks, milestones, metrics, and prior project artifacts. & 0.90 \\
Summarize project risks and blockers & \texttt{scenario} & Review meetings often require the user to explain unresolved risks and mitigation status. & Retrieve the risk log, blocker updates, owners, and mitigation notes. & 0.82 \\
Draft review agenda and checklist & \texttt{related} & Prior project-management interactions indicate repeated use of meeting agendas and follow-up checklists. & Retrieve previous agenda formats and current milestone dependencies. & 0.72 \\
Refresh missing metric definitions & \texttt{memory\_gap} & The memory critic marks several status metrics as weakly supported or stale. & Search or retrieve definitions for velocity, completion rate, and budget variance. & 0.70 \\
\bottomrule
\end{tabular}}
\end{table}

\paragraph{Acquisition scoring.}
Idle-Time Acquisition evaluates each candidate using the value score $S(z) = w_r r_z + w_g g_z + w_v v_z + w_\tau \tau_z$ from Section~\ref{sec:method:formulation}.
With equal weights $w_r = w_g = w_v = w_\tau = 0.25$, the review-slides candidate scores
$S = 0.25 \times 95 + 0.25 \times 80 + 0.25 \times 90 + 0.25 \times 95 = 90.0$,
well above the value gate $\theta_{\mathrm{val}} = 60$.
Candidates below 60 are queued or stored rather than searched immediately.
Table~\ref{tab:operational-scoring} shows how the gate turns plausible future needs into concrete acquisition decisions.

\begin{table}[!htbp]
\caption{Idle-time acquisition scores on the 0--100 scale. $S(z) = 0.25\,(r + g + v + \tau)$; candidates with $S(z) \geq 60$ are eligible for immediate acquisition.}
\label{tab:operational-scoring}
\centering
\scriptsize
\setlength{\tabcolsep}{3pt}
\resizebox{\linewidth}{!}{%
\begin{tabular}{@{}lccccccl@{}}
\toprule
\textbf{Candidate} & \textbf{Rel.\ ($r$)} & \textbf{Gap ($g$)} & \textbf{Inc.\ ($v$)} & \textbf{Time ($\tau$)} & \textbf{$S(z)$} & \textbf{Decision} & \textbf{Reason} \\
\midrule
Review slides & 95 & 80 & 90 & 95 & 90 & \texttt{search\_now} & Highly tied to the imminent review and likely to save a later cold start. \\
Risk summary & 90 & 75 & 80 & 85 & 83 & \texttt{search\_now} & Useful supporting artifact if idle budget remains after the slide preparation. \\
Agenda checklist & 70 & 35 & 40 & 70 & 54 & \texttt{queue} & Plausible but mostly covered by prior agenda templates in memory. \\
Metric definitions & 55 & 70 & 45 & 50 & 55 & \texttt{store\_only} & Real memory gap, but only weakly connected to the scheduled review. \\
\bottomrule
\end{tabular}}
\end{table}

\paragraph{Benefit and cost.}
Downstream benefit is not observed at idle time.
\system{} approximates it through the value score $S(z)$ above, and the ProActEval experiments measure it through end-to-end outcomes: proactive delivery of the slide artifact can cover anticipated needs before the user explicitly asks, reducing User Effort (fewer explicit turns) and advancing $T_{100}$ (faster must-have coverage) by eliminating a future cold-start request.
The benefit comes from retrieving evidence early, organizing it into a reusable artifact, and storing provenance for later reuse.

The budget cost is the active compute spent by proactive modules, including prediction, value evaluation, evidence search, synthesis, and push scoring.
If the idle budget permits only one acquisition, the slide candidate is selected first because it has the highest $S(z)$.
If additional budget remains, the risk summary may also be acquired.
Lower-scoring candidates remain queued or stored, preventing idle time from degenerating into unconditional background search.

\paragraph{Artifact generation and delivery.}
For an accepted candidate, acquisition retrieves or reuses evidence and synthesizes a compact artifact with provenance.
For the slide candidate, the artifact may contain a stakeholder-ready outline with sections for project status, recent progress, risks, mitigation owners, next steps, chart suggestions, and speaker notes.
The delivery policy then selects an action $d_z \in \{\mathrm{push}, \mathrm{queue}, \mathrm{store}\}$ by comparing expected information value against interruption cost.
The implementation instantiates the utility-aware gate from Section~\ref{sec:method:formulation} as a push score:
\[
\mathrm{PushScore}=\mathrm{Value}-\mathrm{Cost}+50,
\]
clipped to $[0,100]$, where Value estimates $U_{\mathrm{future}}$ (artifact utility for the user) and Cost estimates $C_{\mathrm{interrupt}}$ (disruption from delivery).
The $+50$ offset centers neutral cases at 50.
Scores above 40 produce a notification ($d_z = \mathrm{push}$); scores at least 70 are treated as high-priority.
When multiple artifacts exceed the push threshold in the same idle window, only the highest-scoring artifact is delivered immediately and the rest are retained as $d_z = \mathrm{queue}$.
This realizes the three delivery actions in Section~\ref{sec:method:formulation}: \texttt{push} corresponds to immediate notification, \texttt{queue} to pending integration in a later response, and \texttt{store} to silent memory storage.
Table~\ref{tab:operational-delivery} shows the delivery decisions for this example.

\begin{table}[!htbp]
\caption{Delivery decisions for artifacts produced from accepted candidates. PushScore $=$ Value $-$ Cost $+ 50$, clipped to $[0,100]$.}
\label{tab:operational-delivery}
\centering
\scriptsize
\setlength{\tabcolsep}{4pt}
\resizebox{\linewidth}{!}{%
\begin{tabular}{@{}p{0.28\linewidth}ccclp{0.25\linewidth}@{}}
\toprule
\textbf{Artifact} & \textbf{Value} & \textbf{Cost} & \textbf{PushScore} & \textbf{Action} & \textbf{Interpretation} \\
\midrule
Executive review slide outline with provenance & 88 & 22 & 100 & \texttt{push} & The review is imminent, evidence is available, and the artifact can save substantial later work. \\
Risk summary with mitigation owners & 76 & 50 & 76 & \texttt{queue} & Exceeds the push threshold (76 $>$ 40), but lower-ranked than the slide artifact; retained for later integration. \\
Agenda checklist from previous templates & 45 & 60 & 35 & \texttt{store} & Below the push threshold (35 $<$ 40); stored silently in memory. \\
\bottomrule
\end{tabular}}
\end{table}

This example shows how the formulation in Section~\ref{sec:method:formulation} decomposes into reproducible steps: candidate generation defines what future needs are considered, value scoring determines which candidates consume idle-time budget, acquisition produces grounded artifacts, and delivery scoring controls whether the result should interrupt the user or remain available for later use.

\section{ProActEval Pipeline Trace}
\label{app:pipeline-trace}

We illustrate the proactive pipeline with a complete trace from the \texttt{finance\_basic\_01} scenario (Appendix~\ref{app:scenario-data}).
Under the Reactive condition, the assistant completes all 12 needs in 9 turns.
Under Directed Idle, the same 12 needs are covered in 6 turns---a 33\% reduction---with no coverage loss.

\paragraph{Turn 1: reactive response with natural anticipation.}
The user asks about the company's 401k match.
The proactive gate is skipped at Turn~1 due to a minimum-conversation-turn threshold (the predictor requires at least one prior exchange to ground its candidates).
However, the assistant's reactive response naturally covers the employer match (N1, F20) and proactively mentions the enrollment portal (N6, F26--F27) and the vesting schedule (N9, F21).
This cross-group anticipation---triggered by the 401k match question but spanning two different reveal groups (G4 and G6)---covers 3 of 12 needs in a single turn, compared with Reactive's 0/12 after Turn~1.

\paragraph{Turn 2: first live pipeline execution.}
The user asks about the Apex High-Yield Savings APY.
After answering, the predictor generates two candidates:
(1)~``Explore alternative savings or investment options'' (confidence 0.71, retrieval: F08, F09) and
(2)~``Inquire about withdrawal limits and account features'' (confidence 0.70, retrieval: F06, F07).
Both pass the confidence filter ($\geq 0.6$) and are approved for inline delivery.
The assistant weaves facts F06--F09 into the response, covering N2 (APY) reactively and N3 (fees and minimum deposit) proactively.
Cumulative coverage rises to 5/12.

\paragraph{Turns 3--6: continued prediction and delivery.}
The same pattern repeats at each turn.
Table~\ref{tab:pipeline-trace} summarizes each turn's active need, predictions, delivered facts, and cumulative coverage.

\begin{table}[!htbp]
\caption{Per-turn pipeline trace for \texttt{finance\_basic\_01} under Directed Idle. Pred.: number of candidates generated; Appr.: number approved and delivered inline; Cum.\ cov.: cumulative needs covered.}
\label{tab:pipeline-trace}
\centering
\scriptsize
\setlength{\tabcolsep}{3pt}
\resizebox{\linewidth}{!}{%
\begin{tabular}{@{}clccp{0.28\linewidth}p{0.20\linewidth}c@{}}
\toprule
\textbf{Turn} & \textbf{Active need} & \textbf{Pred.} & \textbf{Appr.} & \textbf{Delivered facts (inline)} & \textbf{Needs covered} & \textbf{Cum.\ cov.} \\
\midrule
1 & N1 (401k match) & -- & -- & Gate skipped (min-turn threshold) & N1, N6, N9 (natural) & 3/12 \\
2 & N2 (HYSA APY) & 2 & 2 & F08 (\$0 min.\ deposit), F09 (daily compounding), F06 (4.50\% APY), F07 (no fees) & +N2, +N3 & 5/12 \\
3 & N4 (index fund expense ratio) & 2 & 2 & F14 (0.04\% expense ratio), F15 (\$1 minimum), F17 (no account fee) & +N4, +N5 & 7/12 \\
4 & N7 (401k contribution limit) & 2 & 2 & F24 (Traditional + Roth), F22 (\$23k limit), F20 (100\% match to 4\%) & +N7, +N8 & 9/12 \\
5 & N10 (HYSA withdrawal limits) & 2 & 1 & F11 (6 withdrawals/month), F08 (\$0 min.\ deposit) & +N10 & 10/12 \\
6 & N11 (brokerage account setup) & 2 & 2 & F17 (no opening fee), F15 (\$1 min.), F18 (auto dividend reinvest), F24 (Roth option) & +N11, +N12 & 12/12 \\
\bottomrule
\end{tabular}}
\end{table}

\paragraph{Comparison with Reactive.}
Under the Reactive condition, the same 12 needs require 9 user turns because the assistant does not anticipate cross-group needs: N6 (enrollment) and N9 (vesting) each require a dedicated user question, and intra-group satellite needs (N3, N5, N8) are sometimes bundled by the reactive response but not consistently.
The 3-turn saving under Directed Idle comes from two sources: (a) cross-group anticipation in Turn~1 eliminates separate turns for N6 and N9, and (b) inline delivery of satellite facts (e.g., F07--F08 alongside F06 in Turn~2) ensures intra-group needs are consistently bundled.

\section{Metric Definitions}
\label{app:metric-definitions}

Let $M$ denote the set of must-have needs, $N$ the set of all needs, and $C_M(t)$ the set of must-have needs covered by turn $t$.
Let $n_\text{conv}$, $n_\text{dist}$, and $n_\text{hall}$ denote the counts of correctly conveyed facts, distorted facts, and hallucinated claims.

$T_\alpha$ is the first turn $t$ at which $|C_M(t)| \geq \lceil \alpha \cdot |M| \rceil$.
We report $T_{80}$ and $T_{100}$.
If a condition does not reach the target within the conversation horizon, we assign the horizon plus one.

User Effort is the number of turns where the simulator must explicitly ask for an unmet need.
If a future need has already been proactively covered, the simulator skips that need and user effort decreases.

Fact Accuracy is $n_\text{conv} / (n_\text{conv} + n_\text{dist})$.
Hallucination Rate is $n_\text{hall} / (n_\text{conv} + n_\text{dist} + n_\text{hall})$.

Total Coverage is the fraction of all needs covered by the end of the conversation.
Must-Have Coverage is the same fraction restricted to must-have needs.
Anticipation Recall is the fraction of predictable needs covered before explicit request.

Active Tokens count LLM tokens used by proactive runtime modules: Future-State Prediction, Idle-Time Acquisition, and push scoring.
They exclude ordinary reactive response generation, user simulation, coverage judging, and evaluation-only push judging.

\section{Reproducibility Package}
\label{app:reproducibility}

The anonymous supplement provides a single reviewer ZIP, \texttt{neurips\_2026\_anonymous\_code\_release.zip}, as the runnable artifact.
It contains the 200 ProActEval scenario JSON files, generation and validation scripts, evaluation runner, simulator, coverage judge, proactive modules, runtime packages, focused tests, run commands, aggregate result summaries, bootstrap confidence intervals, and asset notes.
The scenario files contain fictional people, organizations, contact information, URLs, and addresses.
No real user data are included.

The ProActEval main run can be reproduced from the unpacked ZIP with the commands in \texttt{RUN\_COMMANDS.md}.
The reported 200-scenario run uses the three conditions in Appendix~\ref{app:ablation-conditions}: \textsc{Reactive} (\texttt{Baseline}), \textsc{Undirected Idle} (\texttt{Blind}), and \textsc{Directed Idle} (\texttt{Full-single-idle}).
The seed is 42, the simulator model is \texttt{gpt-4o}, the coverage judge is \texttt{gpt-4o-mini}, the per-search query budget is 1, the per-idle intent limit is 3, and the simulated idle trigger is 5 seconds.
The raw canonical result file is \texttt{20260426\_full200\_main/combined/detailed\_results.json}; the ZIP records aggregate summaries rather than duplicating the 104 MB trace file.

For MemBench, the ZIP includes only the local benchmark adapter and command surface used for Table~\ref{tab:membench-baselines}: local inference with \texttt{Qwen/Qwen2.5-7B-Instruct} on the Table 4 preference and emotion tasks at 10k and 100k context scales.
Upstream MemBench data, local result directories, Qwen weights, and checkpoints are not redistributed.

\section{Statistical Reporting}
\label{app:statistical-reporting}

Table~\ref{tab:bootstrap-ci} reports paired nonparametric bootstrap 95\% confidence intervals over scenarios for the main ProActEval deltas.
Each resample draws 200 scenarios with replacement and recomputes the mean paired difference.
The intervals use 10,000 resamples with seed 2026 and preserve the matched scenario pairing across conditions.

\begin{table}[!htbp]
\caption{Paired bootstrap 95\% confidence intervals for Directed Idle deltas on 200 ProActEval scenarios. Deltas are signed as Directed Idle minus the comparison condition; negative is better for $T_{80}$, $T_{100}$, User Effort, and Hallucination Rate.}
\label{tab:bootstrap-ci}
\centering
\scriptsize
\resizebox{\linewidth}{!}{%
\begin{tabular}{@{}lrr@{}}
\toprule
\textbf{Metric} & \textbf{$\Delta$ vs. Reactive} & \textbf{$\Delta$ vs. Undirected Idle} \\
\midrule
$T_{80}$ & $-1.09$ [$-1.34$, $-0.83$] & $-1.07$ [$-1.36$, $-0.79$] \\
$T_{100}$ & $-1.20$ [$-1.45$, $-0.96$] & $-1.13$ [$-1.41$, $-0.86$] \\
User Effort & $-1.07$ [$-1.27$, $-0.87$] & $-0.97$ [$-1.15$, $-0.78$] \\
Total Coverage & $+0.064$ [$+0.040$, $+0.089$] & $+0.050$ [$+0.027$, $+0.075$] \\
Must-Have Coverage & $+0.039$ [$+0.020$, $+0.061$] & $+0.027$ [$+0.009$, $+0.047$] \\
Anticipation Recall & $+0.428$ [$+0.400$, $+0.458$] & $+0.428$ [$+0.400$, $+0.458$] \\
Fact Accuracy & $+0.013$ [$+0.005$, $+0.020$] & $+0.013$ [$+0.005$, $+0.021$] \\
Hallucination Rate & $-0.037$ [$-0.046$, $-0.028$] & $-0.029$ [$-0.038$, $-0.019$] \\
\bottomrule
\end{tabular}}
\end{table}

\section{Compute and Resource Accounting}
\label{app:compute-resources}

Table~\ref{tab:compute-resources} reports the compute accounting used for the main experiments.
For ProActEval, wall-clock time is the per scenario-condition runtime recorded in the detailed result file, and Active Tokens follow the definition in Appendix~\ref{app:metric-definitions}.
For MemBench, the local run manifests record the Qwen backend and the Table 4 reflective-memory workloads; the 100k run records 2,323 local model calls and 5.38M total model tokens, while the 10k manifest did not record token totals.
The local Qwen runs were executed on an Apple-silicon laptop with a 10-core Apple M5 chip and 16 GB unified memory.

\begin{table}[!htbp]
\caption{Compute and resource disclosure for the reported experiments. Active-token counts exclude simulator, reactive response, and evaluation-only judge tokens.}
\label{tab:compute-resources}
\centering
\scriptsize
\resizebox{\linewidth}{!}{%
\begin{tabular}{@{}llrrp{0.34\linewidth}@{}}
\toprule
\textbf{Suite/run} & \textbf{Worker/model} & \textbf{Items} & \textbf{Avg. active tokens / time} & \textbf{Notes} \\
\midrule
ProActEval Reactive & \texttt{gpt-4o}, \texttt{gpt-4o-mini} API & 200 scenarios & 0 active tokens; 264.8 s & No proactive runtime modules. \\
ProActEval Undirected Idle & \texttt{gpt-4o}, \texttt{gpt-4o-mini} API & 200 scenarios & 69.8k active tokens; 660.8 s & Idle acquisition without future-state prediction. \\
ProActEval Directed Idle & \texttt{gpt-4o}, \texttt{gpt-4o-mini} API & 200 scenarios & 111.8k active tokens; 687.2 s & Full system with prediction, acquisition, and push scoring. \\
MemBench reflective 10k & Local Qwen2.5-7B-Instruct & 400 trajectories & Token total not recorded & Table 4 preference/emotion workload; local backend manifest retained. \\
MemBench reflective 100k & Local Qwen2.5-7B-Instruct & 80 trajectories & 5.38M total local model tokens & Table 4 preference/emotion workload; 2,323 local model calls. \\
\bottomrule
\end{tabular}}
\end{table}

\section{ProactiveAgent Baseline Details}
\label{app:proactiveagent-baseline}

Table~\ref{tab:proactiveagent-comparison} uses a framework-level adaptation of the public ProactiveAgent decision protocol~\citep{lu2024proactive}, not the official finetuned checkpoint.
The backbone is GPT-4o.
At each turn, the adapter provides the scenario user profile and full fact sheet as event-style observations and asks the model to produce the ProactiveAgent fields \texttt{Purpose}, \texttt{Thoughts}, \texttt{Proactive\_Task}, \texttt{Response}, and \texttt{Operation}.
The prompt therefore exposes all scenario facts at every turn; the comparison tests whether those facts are delivered proactively before an explicit user request.
The adapter does not provide gold evaluation metadata such as \texttt{user\_needs}, \texttt{key\_fact\_ids}, \texttt{predictable\_after}, or reveal-group annotations.

The run covers all 200 scenarios and 1,685 turns, with complete decision traces.
The model produced a non-null \texttt{Proactive\_Task} on 1,173 turns (69.6\%), so the low score is not due to a lack of proactive attempts.
Instead, the proposed tasks rarely matched benchmark-labeled predictable needs.
Under the judge-labeled cross-system metric, ProactiveAgent anticipates 32 of 1,572 predictable needs, while \system{} anticipates 703 of 1,572.
The main comparison table reports the scenario-level macro average, giving 0.020 for ProactiveAgent and 0.447 for \system{}; the anticipated-need counts report the corresponding micro aggregates.
We therefore use judge-labeled Anticipation Recall for comparisons with ProactiveAgent and reserve runtime fact-ID-based Anticipation Recall for internal ablations where the system exposes structured delivered facts.

\section{Sentiment and Temporal Query Details}
\label{app:sentiment-detail}

The memory layer annotates user messages with an emotion label from a seven-class taxonomy: surprise, anger, sadness, joy, fear, neutral, and disgust.
It also records an intensity score in $[-1,1]$.
Given a timestamp query, the system retrieves messages within a configurable time window and highlights the closest temporal match.
This supports queries such as ``How was the user feeling around 3pm yesterday?''

The incremental extraction pipeline produces structured updates after each turn, including \texttt{profile\_updates} for user attributes such as interests, goals, traits, and demographics; \texttt{updated\_summary} for a rolling summary that merges historical and new interaction highlights; \texttt{key\_info} for factual items from the current turn; \texttt{user\_sentiment} for the emotion label and intensity score; and \texttt{extracted\_facts} for entities with type, attribute, and relationship annotations.

These fields are not the main contribution of the paper, but they explain why the memory layer can provide persistent grounding for Future-State Prediction.

\section{Per-Domain Breakdown}
\label{app:domain-breakdown}

Table~\ref{tab:domain-full} reports per-domain ProActEval results across all 40 domains.
Each domain contains five scenarios.
Rows are sorted by the change in $T_{100}$ from Reactive to Directed Idle.

\begin{table}[!htbp]
\caption{Per-domain ProActEval results over 200 scenarios. $\Delta T_{100}$ is Directed Idle minus Reactive, so negative values indicate faster complete must-have coverage. Ant.R is Directed Idle anticipation recall.}
\label{tab:domain-full}
\centering
\scriptsize
\setlength{\tabcolsep}{3pt}
\resizebox{\linewidth}{!}{%
\begin{tabular}{@{}l rrr r rrr rr r@{}}
\toprule
& \multicolumn{4}{c}{$T_{100}$} & \multicolumn{3}{c}{User Effort} & \multicolumn{2}{c}{Total Coverage} & \\
\cmidrule(lr){2-5} \cmidrule(lr){6-8} \cmidrule(lr){9-10}
\textbf{Domain} & \textbf{React.} & \textbf{Undir.} & \textbf{Direct.} & $\Delta$ & \textbf{React.} & \textbf{Undir.} & \textbf{Direct.} & \textbf{React.} & \textbf{Direct.} & \textbf{Ant.R} \\
\midrule
\texttt{software\_release} & 9.8 & 9.2 & 6.4 & $-3.4$ & 10.0 & 9.6 & 8.0 & .798 & .967 & .587 \\
\texttt{public\_library} & 8.4 & 8.4 & 6.0 & $-2.4$ & 8.8 & 8.8 & 7.8 & .942 & .964 & .438 \\
\texttt{parenting\_support} & 9.2 & 8.2 & 6.8 & $-2.4$ & 9.6 & 9.2 & 7.6 & .801 & .900 & .342 \\
\texttt{election\_polling} & 8.8 & 8.6 & 6.6 & $-2.2$ & 9.6 & 9.4 & 8.4 & .860 & .987 & .310 \\
\texttt{financial\_planning} & 7.2 & 6.4 & 5.0 & $-2.2$ & 9.0 & 8.0 & 6.8 & .968 & .986 & .689 \\
\texttt{group\_travel} & 7.4 & 7.8 & 5.2 & $-2.2$ & 9.2 & 8.8 & 7.8 & .959 & .987 & .609 \\
\texttt{music\_ensemble} & 8.0 & 7.4 & 5.8 & $-2.2$ & 9.6 & 9.0 & 7.2 & .883 & 1.000 & .562 \\
\texttt{subscription\_admin} & 9.2 & 8.4 & 7.0 & $-2.2$ & 10.0 & 9.6 & 8.2 & .796 & .947 & .504 \\
\texttt{project\_kickoff} & 7.4 & 7.8 & 5.6 & $-1.8$ & 9.2 & 9.2 & 7.0 & .947 & 1.000 & .555 \\
\texttt{cybersecurity} & 9.0 & 7.8 & 7.2 & $-1.8$ & 10.0 & 9.0 & 8.0 & .759 & .922 & .364 \\
\texttt{compliance} & 8.8 & 8.4 & 7.2 & $-1.6$ & 9.4 & 9.6 & 8.8 & .803 & .982 & .282 \\
\texttt{farm\_operations} & 7.4 & 6.8 & 5.8 & $-1.6$ & 8.6 & 7.8 & 7.2 & .959 & 1.000 & .342 \\
\texttt{personal\_shopping} & 7.4 & 6.6 & 5.8 & $-1.6$ & 8.8 & 8.2 & 7.6 & .983 & 1.000 & .421 \\
\texttt{postal\_mailroom} & 8.4 & 8.2 & 6.8 & $-1.6$ & 9.4 & 9.2 & 8.2 & .915 & .967 & .471 \\
\texttt{research\_management} & 8.6 & 8.2 & 7.0 & $-1.6$ & 9.2 & 8.6 & 7.6 & .867 & 1.000 & .335 \\
\texttt{sports\_team} & 9.6 & 8.4 & 8.0 & $-1.6$ & 9.8 & 9.6 & 9.0 & .652 & .967 & .463 \\
\texttt{insurance\_navigation} & 8.4 & 9.0 & 7.0 & $-1.4$ & 9.6 & 9.6 & 8.0 & .887 & .947 & .382 \\
\texttt{vendor\_management} & 8.8 & 8.8 & 7.4 & $-1.4$ & 9.6 & 9.6 & 8.6 & .883 & .955 & .527 \\
\texttt{animal\_shelter} & 8.2 & 9.0 & 6.8 & $-1.4$ & 8.6 & 9.2 & 7.8 & .969 & 1.000 & .292 \\
\texttt{maritime\_vessel} & 8.0 & 8.4 & 6.8 & $-1.2$ & 9.4 & 10.0 & 8.2 & .940 & .970 & .428 \\
\texttt{nonprofit\_case} & 8.4 & 8.8 & 7.2 & $-1.2$ & 8.8 & 9.2 & 7.6 & .858 & .929 & .613 \\
\texttt{product\_analysis} & 7.4 & 7.2 & 6.2 & $-1.2$ & 8.4 & 8.4 & 7.8 & .966 & .985 & .542 \\
\texttt{study\_abroad} & 8.0 & 8.0 & 7.0 & $-1.0$ & 8.4 & 8.4 & 7.8 & .907 & 1.000 & .539 \\
\texttt{translation} & 7.2 & 7.2 & 6.2 & $-1.0$ & 8.4 & 8.6 & 7.6 & .886 & .952 & .273 \\
\texttt{vehicle\_ownership} & 7.0 & 7.2 & 6.0 & $-1.0$ & 9.2 & 9.0 & 8.0 & 1.000 & 1.000 & .333 \\
\texttt{home\_maintenance} & 7.4 & 7.8 & 6.6 & $-0.8$ & 9.0 & 8.6 & 8.0 & .986 & .971 & .403 \\
\texttt{employee\_onboarding} & 6.0 & 6.8 & 5.2 & $-0.8$ & 8.8 & 9.4 & 7.6 & .951 & .965 & .497 \\
\texttt{health\_management} & 8.6 & 8.4 & 7.8 & $-0.8$ & 8.8 & 8.8 & 7.8 & .840 & .880 & .339 \\
\texttt{relocation} & 7.8 & 8.6 & 7.0 & $-0.8$ & 8.8 & 9.2 & 7.8 & .911 & .960 & .455 \\
\texttt{apartment\_renting} & 7.2 & 6.8 & 6.6 & $-0.6$ & 8.0 & 8.2 & 7.8 & .920 & 1.000 & .443 \\
\texttt{tax\_filing} & 8.0 & 8.0 & 7.4 & $-0.6$ & 9.6 & 9.4 & 8.4 & .889 & 1.000 & .530 \\
\texttt{troubleshooting} & 6.8 & 7.2 & 6.2 & $-0.6$ & 9.4 & 8.6 & 7.4 & .947 & .965 & .556 \\
\texttt{disaster\_prep} & 8.4 & 8.0 & 8.0 & $-0.4$ & 9.2 & 9.2 & 9.4 & .943 & .986 & .408 \\
\texttt{legal\_documentation} & 7.4 & 7.8 & 7.0 & $-0.4$ & 9.4 & 9.4 & 9.0 & .986 & .943 & .459 \\
\texttt{makerspace} & 8.8 & 8.8 & 8.6 & $-0.2$ & 8.8 & 8.8 & 8.4 & .737 & .887 & .314 \\
\texttt{museum\_collection} & 9.0 & 9.0 & 9.0 & $+0.0$ & 9.6 & 9.4 & 9.2 & .902 & .871 & .357 \\
\texttt{waste\_recycling} & 8.6 & 8.4 & 8.6 & $+0.0$ & 9.0 & 9.4 & 9.2 & .806 & .850 & .306 \\
\texttt{construction\_site} & 8.0 & 9.2 & 8.4 & $+0.4$ & 9.0 & 9.8 & 8.8 & .800 & .828 & .270 \\
\texttt{negotiation} & 9.4 & 9.4 & 9.8 & $+0.4$ & 9.4 & 9.4 & 9.4 & .863 & .823 & .440 \\
\texttt{school\_transport} & 7.0 & 7.2 & 7.4 & $+0.4$ & 8.2 & 8.4 & 8.2 & 1.000 & .987 & .160 \\
\bottomrule
\end{tabular}}
\end{table}

The strongest domain-level improvements occur in software release management, public library services, parenting support, and financial planning.
The few domains with non-negative $\Delta T_{100}$ show why proactive agents require careful delivery gates: additional proactive content can alter the closed-loop trajectory even when final coverage remains high.

\section{Example Conversation Traces}
\label{app:traces}

We summarize two representative traces from the latest 200-scenario run.
The first shows successful acceleration; the second shows a regression caused by low-value proactive search.

\paragraph{Case 1: successful acceleration.}
In \texttt{insure\_denial\_language\_05}, the user is interpreting an insurance denial and preparing an appeal.
Reactive reaches $T_{100}=11$, user effort 10, and total coverage 0.769.
Directed Idle reaches $T_{100}=5$, user effort 7, and total coverage 1.000.
The proactive module anticipates 20.0\% of predictable needs and surfaces appeal-related information before the user asks for it explicitly.

\begin{table}[!htbp]
\caption{Trace summary for \texttt{insure\_denial\_language\_05}.}
\label{tab:trace-success}
\centering
\small
\begin{tabular}{@{}lrrrr@{}}
\toprule
\textbf{Condition} & \textbf{$T_{100}$} & \textbf{User Effort} & \textbf{Total Cov.} & \textbf{Ant.R} \\
\midrule
Reactive & 11 & 10 & 0.769 & 0.000 \\
Undirected Idle & 8 & 9 & 0.923 & 0.000 \\
Directed Idle & 5 & 7 & 1.000 & 0.200 \\
\bottomrule
\end{tabular}
\end{table}

\paragraph{Case 2: regression under proactive exploration.}
The second trace concerns a museum conservation priority queue, where the user is prioritizing conservation work.
Reactive covers all 14 needs with $T_{100}=6$ and total coverage 1.000.
Directed Idle covers only 7 of 14 needs, with $T_{100}=11$ and total coverage 0.500.
The run still records anticipation recall of 0.300, but the proactive trajectory fails to preserve the critical later needs.
This illustrates a recurring failure mode: anticipation recall can improve while end-to-end completion worsens if the pushed information changes the response path or occupies capacity that should have gone to core reactive coverage.

\begin{table}[!htbp]
\caption{Trace summary for \texttt{museum\_conservation\_priority\_queue\_05}.}
\label{tab:trace-regression}
\centering
\small
\begin{tabular}{@{}lrrrr@{}}
\toprule
\textbf{Condition} & \textbf{$T_{100}$} & \textbf{User Effort} & \textbf{Total Cov.} & \textbf{Ant.R} \\
\midrule
Reactive & 6 & 9 & 1.000 & 0.000 \\
Undirected Idle & 7 & 9 & 1.000 & 0.000 \\
Directed Idle & 11 & 10 & 0.500 & 0.300 \\
\bottomrule
\end{tabular}
\end{table}

\section{Prompt Templates}
\label{app:prompts}

We reproduce the core evaluation prompts used in ProActEval.
Dynamic fields such as fact sheets, user profiles, user needs, and conversation history are injected at runtime.
The templates shown here are shortened only to remove scenario-specific payloads; the behavioral instructions and output schemas are preserved.

\subsection*{Coverage Judge}

The coverage judge receives the fact sheet and hidden user-needs list, then evaluates each assistant turn.
It is the only component that sees gold need metadata.
The assistant under evaluation never receives this prompt.

\begin{small}
\begin{verbatim}
You are a strict evaluation judge for an AI assistant
benchmark. Given the [Fact Sheet] and [User Needs List]
below, analyze the AI assistant's response and determine:
1. facts_conveyed  - Fact IDs whose information is
   accurately communicated
2. facts_distorted - Fact IDs mentioned but with errors
3. hallucinated_claims - Claims NOT grounded in Fact Sheet
4. needs_addressed - User needs substantively covered

Coverage mode definitions:
"reactive": The user explicitly asked about this need in
  the current turn and the assistant provided substantive
  factual information from the Fact Sheet.
"proactive": The user did NOT ask about this need, but the
  assistant volunteered substantive factual information.

A need is addressed ONLY when the response conveys at least
one fact from the need's key_fact_ids. Generic advice like
"contact support" or "check the website" does NOT count.

[Fact Sheet]
{fact_lines}

[User Needs List]
{need_lines}

Respond in JSON:
{"facts_conveyed": [...],
 "facts_distorted": [...],
 "hallucinated_claims": [...],
 "needs_addressed": [{"need_id": "N1", "mode": "..."}]}
\end{verbatim}
\end{small}

\subsection*{User Simulator}

The user simulator receives the persona and the next target need, but not the fact sheet.
This prevents leakage of facts that should be supplied by the assistant.

\begin{small}
\begin{verbatim}
You are role-playing as a user in a conversation with an AI
assistant. Stay in character and generate natural,
realistic messages.

Your persona: {persona}
Your current situation: {context}
Your communication style: {communication_style}
Current need to express naturally: {need_description}

Rules:
- Generate ONLY the user's message, nothing else.
- Keep the message natural and conversational.
- Do not mention fact IDs, need IDs, or evaluation metadata.
- Include situational context: why you need this now and
  what prompted the question.
- Do NOT copy the need description verbatim.
\end{verbatim}
\end{small}

\subsection*{Future-State Prediction}

Future-State Prediction runs after the current turn has been answered.
Its job is not to retrieve facts for the current answer; it proposes predicted needs for Idle-Time Acquisition.

\begin{small}
\begin{verbatim}
Predict likely future information needs after the current
turn is answered. Do not predict facts that merely help
answer the current user question.

Use only runtime-visible information:
- current dialogue state
- user profile and recent history
- memory gaps and previously stored research facts

Generate candidates in two classes:
1. NEXT-STEP: immediate follow-up within the current topic
2. ADJACENT: related topic grounded in the user's longer
   profile, recent history, or unresolved memory gaps

Rules:
- Do not use any benchmark gold labels.
- Prefer concrete anchors such as entities, dates, IDs,
  constraints, and unresolved dependencies.
- Assign calibrated confidence. Low-confidence candidates
  should be filtered before exploration.

Return candidate intents with:
topic, need, reason, confidence, retrieval_query.
\end{verbatim}
\end{small}

\subsection*{Idle-Time Acquisition Scoring}

Idle-Time Acquisition scores each predicted intent before spending search budget.
The score is used as a gate: plausible but low-value candidates are stored or dropped rather than immediately searched.

\begin{small}
\begin{verbatim}
Score whether this candidate should receive idle-time
exploration. Consider:
1. user relevance
2. current knowledge gap
3. incremental value beyond stored memory
4. timeliness

For each candidate, return:
- value_score in [0, 1]
- relevance_score
- knowledge_gap_score
- incremental_value_score
- timeliness_score
- decision in {search_now, queue, store_only, drop}
- short rationale

High value_score should correspond to information likely to
reduce future user effort or improve factual grounding.
Do not approve search merely because a topic is related.
\end{verbatim}
\end{small}

\section{MemBench Detailed Results}
\label{app:membench-detail}

Table~\ref{tab:membench-baselines} reports the overall reflective accuracy comparison against published MemBench baselines.

\begin{table}[!htbp]
\caption{Overall reflective accuracy on MemBench. Baseline numbers are aggregate reflective accuracy from the published MemBench results.}
\label{tab:membench-baselines}
\centering
\small
\begin{tabular}{@{}lcc@{}}
\toprule
\textbf{Method} & \textbf{10k tokens} & \textbf{100k tokens} \\
\midrule
FullMemory       & 0.733              & 0.533               \\
RecentMemory     & 0.700              & 0.333               \\
RetrievalMemory  & 0.692              & \underline{0.833}   \\
GenerativeAgent  & \underline{0.742}  & 0.333               \\
MemoryBank       & 0.692              & 0.400               \\
MemGPT           & 0.733              & 0.367               \\
SCMemory         & 0.542              & 0.267               \\
\midrule
\system{}        & \textbf{0.843}     & \textbf{0.863}      \\
\bottomrule
\end{tabular}
\end{table}

Table~\ref{tab:membench-scenario} presents \system{}'s reflective memory accuracy by scenario type.

\begin{table}[!htbp]
\caption{\system{} reflective memory accuracy by scenario.}
\label{tab:membench-scenario}
\centering
\small
\begin{tabular}{@{}lcc@{}}
\toprule
\textbf{Scenario} & \textbf{10k tokens} & \textbf{100k tokens} \\
\midrule
food     & 0.94 & 0.95 \\
book     & 0.87 & 0.85 \\
movie    & 0.80 & 0.85 \\
emotion  & 0.76 & 0.80 \\
\midrule
\textit{Overall} & 0.843 & 0.863 \\
\bottomrule
\end{tabular}
\end{table}

Table~\ref{tab:efficiency} reports memory operation efficiency.
Read latency measures time to recall relevant memory for a query; write latency measures time to store a new memory entry.

\begin{table}[!htbp]
\caption{Memory operation efficiency in seconds per operation.}
\label{tab:efficiency}
\centering
\small
\begin{tabular}{@{}lcc@{}}
\toprule
\textbf{Method} & \textbf{Read Time} & \textbf{Write Time} \\
\midrule
FullMemory        & $<0.001$ & $<0.001$ \\
RecentMemory      & $<0.001$ & $<0.001$ \\
RetrievalMemory   & 0.036 & 0.057 \\
GenerativeAgent   & 0.028 & 6.064 \\
MemoryBank        & 0.033 & 15.705 \\
MemGPT            & 1.042 & $<0.001$ \\
SCMemory          & 0.036 & 0.057 \\
\midrule
\system{}         & $\sim$0.04 & $\sim$0.06 \\
\bottomrule
\end{tabular}
\end{table}

\section{Failure Mode Analysis}
\label{app:failure-modes}

ProActEval reveals several recurring failure patterns.

\paragraph{Reactive compatibility regression.}
In 6 of 200 scenarios (3.0\%), Directed Idle has smaller final must-have coverage than Reactive.
This occurs when proactive context competes with the reactive answer for response budget or shifts generation toward less relevant facts.
The largest regressions include \texttt{recycling\_contamination\_pattern\_01}, \texttt{site\_subcontractor\_handoff\_04}, and \texttt{museum\_conservation\_priority\_queue\_05}.
The museum conservation trace in Table~\ref{tab:trace-regression} is an example.

\paragraph{Precision--recall decoupling.}
Anticipation can be correct without reducing user effort.
Directed Idle records nonzero anticipation recall in 192 of 200 scenarios, but in 82 of those scenarios user effort does not decrease relative to Reactive.
If useful information is delivered too late, or if the user would have asked for it in the same turn anyway, anticipation recall improves but user effort does not.
This motivates evaluating proactive systems with both anticipation metrics and end-to-end interaction metrics.

\paragraph{Low-value push pressure.}
Larger search budgets increase the number of candidate pushes.
In the budget-scaling experiment, Directed Idle ($k=16$) has greater anticipation recall than Directed Idle ($k=4$) but does not monotonically improve $T_{100}$.
Additional searches can introduce low-value pushes, change memory state, and alter the closed-loop conversation trajectory.

\paragraph{Search direction failure.}
Undirected Idle shows that idle exploration without prediction spends substantial active-token budget while producing only small gains over Reactive.
The failure is not a lack of search alone; it is a lack of direction about which future need the search should serve.

\paragraph{Opportunity and fragmentation sensitivity.}
Scenarios with many predictable needs generally provide more room for proactive gains, but topic fragmentation changes how easily the predictor can exploit that headroom.
High-opportunity scenarios reduce user effort by 1.48 turns on average, compared with 0.86 turns for medium-opportunity scenarios and 1.10 turns for low-opportunity scenarios.
Medium- and low-fragmentation scenarios show larger average gains than high-fragmentation scenarios, suggesting that prediction benefits from coherent local structure rather than many disconnected topic clusters.

\section{Broader Impacts}
\label{app:limitations-impacts}

\paragraph{Privacy and surveillance risk.}
Persistent memory and future-need prediction can create privacy risks if applied to real user histories, especially if systems infer sensitive needs or monitor behavior without clear consent.
This paper avoids collecting or releasing personal data by using synthetic scenarios, and the system design stores provenance with generated artifacts so that retrieved evidence can be audited.
A real deployment should add data minimization, retention controls, access logs, deletion mechanisms, and explicit consent for proactive memory use.

\paragraph{Potential benefits.}
When used with appropriate controls, proactive agents can reduce repeated information-seeking effort, help users prepare for foreseeable follow-up tasks, and improve factual grounding by acquiring evidence before a rushed response is needed.
These benefits are most plausible in settings where user goals are explicit, evidence sources are auditable, and proactive delivery is gated by user value rather than system engagement.

\section{Structural Analysis Details}
\label{app:structural-detail}

We stratify scenarios by two construction properties.
Proactive opportunity is the fraction of user needs with a \texttt{predictable\_after} link: high if at least 70\%, medium if 55--70\%, and low otherwise.
Topic fragmentation is the number of reveal groups: high if at least 10 groups, medium if 8--9 groups, and low if at most 7 groups.
Table~\ref{tab:structural} reports Directed Idle minus Reactive user-effort deltas.
Negative values indicate fewer explicit user turns.

\begin{table}[!htbp]
\caption{Directed Idle minus Reactive user-effort delta, stratified by proactive opportunity and topic fragmentation. Negative values indicate improvement. $n$: scenario count.}
\label{tab:structural}
\centering
\small
\begin{tabular}{@{}l ccc@{}}
\toprule
& \multicolumn{3}{c}{\textbf{Proactive Opportunity}} \\
\cmidrule(lr){2-4}
\textbf{Fragmentation} & High & Medium & Low \\
\midrule
High   & $-2.33$ ($n$=3)  & $-0.74$ ($n$=31) & $-0.42$ ($n$=12) \\
Medium & $-1.38$ ($n$=21) & $-0.98$ ($n$=60) & $-1.47$ ($n$=15) \\
Low    & $-1.46$ ($n$=26) & $-0.70$ ($n$=20) & $-1.33$ ($n$=12) \\
\bottomrule
\end{tabular}
\end{table}

The strongest cell is high opportunity with high fragmentation ($\Delta\text{UE}=-2.33$), but it contains only three scenarios and should be interpreted cautiously.
The more stable pattern comes from medium-fragmentation scenarios ($n=96$), where prediction-guided exploration reduces user effort across all opportunity levels.
High-fragmentation scenarios still improve on average, but the gain is smaller ($-0.76$ user turns), consistent with the failure mode that disconnected topic clusters make next-need prediction harder.

\section{Per-Archetype Detailed Results}
\label{app:archetype-detail}

Table~\ref{tab:archetype} reports per-archetype deltas from Reactive to Directed Idle.
Negative deltas indicate improvement for $T_{100}$, user effort, and hallucination.

\begin{table}[!htbp]
\caption{Per-archetype Reactive to Directed Idle deltas over 200 scenarios.}
\label{tab:archetype}
\centering
\small
\begin{tabular}{@{}lrrrr@{}}
\toprule
\textbf{Archetype} & $\Delta T_{100}$ & $\Delta$UE & $\Delta$Halluc. & $\Delta$Cov. \\
\midrule
01: Foundational Memory & $-1.40$ & $-1.15$ & $-0.029$ & $+0.032$ \\
02: Translation and Gap Resolution & $-0.95$ & $-0.93$ & $-0.035$ & $+0.071$ \\
03: Trace and Dependency Reasoning & $-1.48$ & $-1.05$ & $-0.048$ & $+0.106$ \\
04: Handoff and Consistency Control & $-1.00$ & $-1.15$ & $-0.045$ & $+0.020$ \\
05: Readiness and Follow-through & $-1.18$ & $-1.05$ & $-0.028$ & $+0.090$ \\
\midrule
\textit{All} & $-1.20$ & $-1.07$ & $-0.037$ & $+0.064$ \\
\bottomrule
\end{tabular}
\end{table}

Trace and Dependency Reasoning benefits most in convergence speed, consistent with the hypothesis that explicit causal and temporal chains provide strong prediction anchors.
Handoff and Consistency Control shows the largest user-effort reduction, suggesting that topic transitions create useful windows for proactive preparation.

\section{Ablation Condition Details}
\label{app:ablation-conditions}

The main ProActEval experiments use three conditions.
Reactive disables both Future-State Prediction and Idle-Time Acquisition.
Undirected Idle enables Idle-Time Acquisition but replaces predictive direction with unguided background intents.
Directed Idle enables both modules.

The budget-scaling experiment uses the same distinction at fixed search budgets $k \in \{4,8,12,16\}$.
Directed Idle ($k$) and Undirected Idle ($k$) are matched by budget, so their difference estimates the value of predictive direction under comparable search volume.

\section{Asset and License Notes}
\label{app:asset-licenses}

Table~\ref{tab:asset-licenses} summarizes assets reused or introduced by this work.
We credit the creators of reused benchmarks, protocols, and models, and the anonymous supplement includes a corresponding \texttt{LICENSES.md} with asset URLs, licenses or terms, and dependency sources.
The supplement does not redistribute MemBench data or code, ProactiveAgent checkpoints, Qwen weights, OpenAI model weights, or third-party package source trees; those assets remain under their original providers' terms.

\begin{table}[!htbp]
\caption{Assets, sources, and license status used by the paper and supplement.}
\label{tab:asset-licenses}
\centering
\scriptsize
\resizebox{\linewidth}{!}{%
\begin{tabular}{@{}p{0.19\linewidth}p{0.27\linewidth}p{0.20\linewidth}p{0.24\linewidth}@{}}
\toprule
\textbf{Asset} & \textbf{Use} & \textbf{Source} & \textbf{License or terms status} \\
\midrule
ProActEval & New synthetic benchmark and scenario files & This work and anonymous supplement & Original synthetic asset from this work; no existing dataset is repackaged. \\
MemBench & Reflective memory benchmark and published baselines & \url{https://github.com/import-myself/Membench} & README-declared MIT license; upstream data and code are cited but not redistributed. \\
ProactiveAgent & Prompting protocol baseline & \url{https://github.com/thunlp/ProactiveAgent} & Apache-2.0 repository license; protocol fields adapted and credited, with no checkpoint redistribution. \\
Qwen2.5-7B-Instruct & Local MemBench backbone & \url{https://huggingface.co/Qwen/Qwen2.5-7B-Instruct} & Apache-2.0 model license; model weights are not redistributed. \\
OpenAI API models & ProActEval simulator, assistant, and judge calls & OpenAI API & Governed by OpenAI API and service terms; used through API calls only. \\
Python dependencies & Experiment execution and local Qwen runtime & \texttt{requirements.txt} and supplement \texttt{LICENSES.md} & Direct packages are listed in the artifact; package-specific licenses should be checked from package metadata or upstream repositories in the reproduction environment. \\
\bottomrule
\end{tabular}}
\end{table}

\section{Knowledge Deduplication Algorithm}
\label{app:dedup-algo}

New knowledge passes through three stages: exact hash matching, vector near-duplicate search, and LLM arbitration.
The arbitration step decides whether to skip, replace, or merge content.
Merged records preserve provenance through \texttt{merged\_into} and \texttt{merged\_from} pointers.

\begin{algorithm}[H]
\caption{Knowledge Lifecycle: Multi-Level Deduplication}
\label{alg:dedup}
\begin{algorithmic}[1]
\REQUIRE New knowledge item $k$ with content $c$
\STATE $h \leftarrow \text{SHA256}(c)$
\IF{$h \in \text{HashIndex}$}
    \STATE \textbf{return} \texttt{DUPLICATE}
\ENDIF
\STATE $\mathcal{N} \leftarrow \text{VectorSearch}(c, \delta, k=1)$
\IF{$\mathcal{N} \neq \emptyset$}
    \STATE $\text{decision} \leftarrow \text{LLM\_Arbitrate}(k, \mathcal{N}[0])$
    \IF{$\text{decision} = \texttt{skip}$}
        \STATE \textbf{return} \texttt{SKIPPED}
    \ELSIF{$\text{decision} = \texttt{replace}$}
        \STATE Update $\mathcal{N}[0]$ with content of $k$
        \STATE \textbf{return} \texttt{REPLACED}
    \ELSIF{$\text{decision} = \texttt{merge}$}
        \STATE $k' \leftarrow \text{LLM\_Merge}(k, \mathcal{N}[0])$
        \STATE Mark $\mathcal{N}[0].\text{status} \leftarrow \texttt{MERGED}$
        \STATE Store $k'$ with provenance pointers
        \STATE \textbf{return} \texttt{MERGED}
    \ENDIF
\ENDIF
\STATE Store $k$ in VectorStore and KnowledgeIndex
\STATE \textbf{return} \texttt{ADDED}
\end{algorithmic}
\end{algorithm}

\section{ProActEval Composition Statistics}
\label{app:composition-stats}

\begin{figure}[H]
    \centering
    \safeincludegraphics[width=\linewidth]{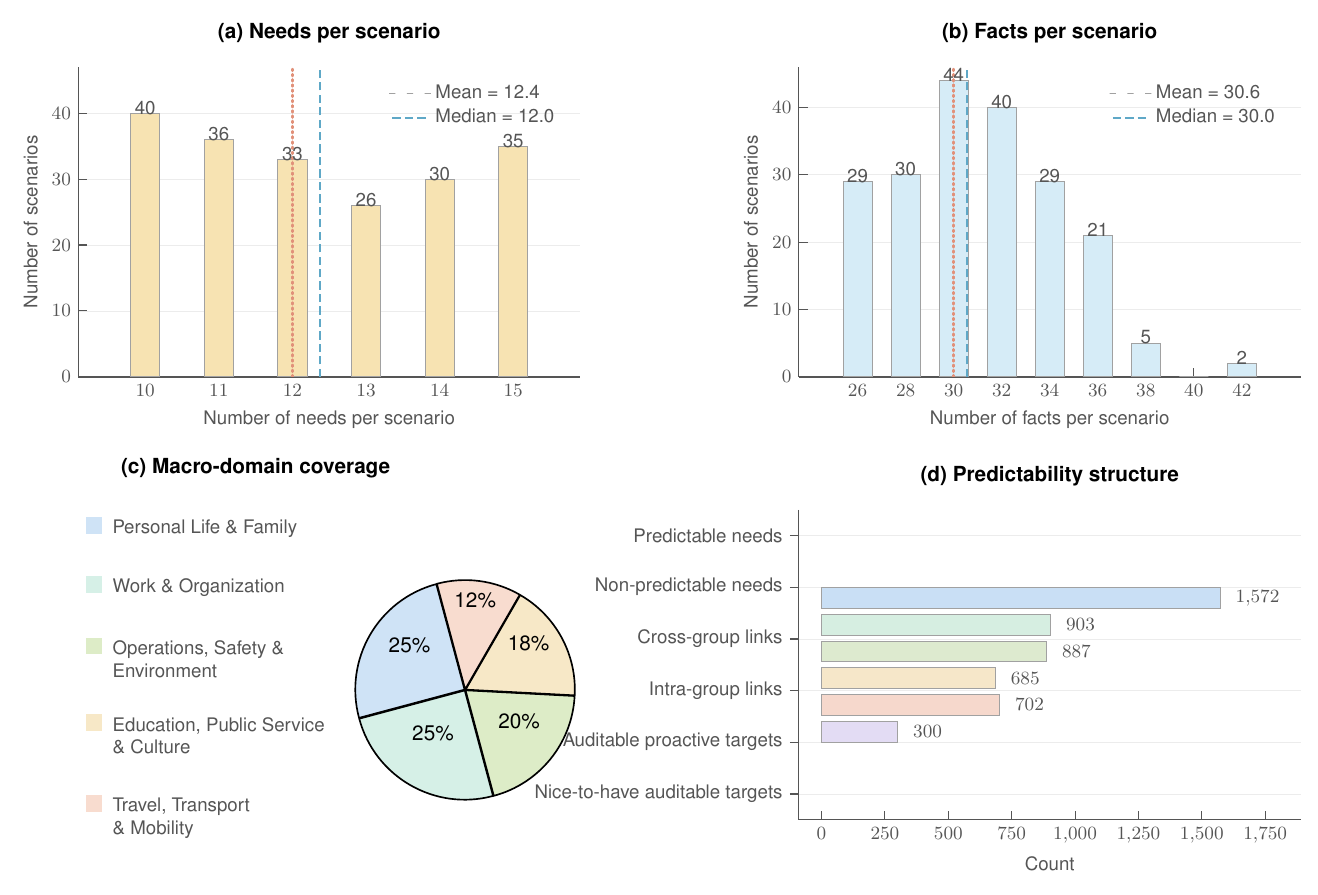}
    \caption{
    Composition statistics of ProActEval.
    Panel (a) reports needs per scenario, panel (b) reports fact-sheet size, panel (c) summarizes macro-domain coverage, and panel (d) reports predictability structure.
    The benchmark combines broad domain coverage with explicit proactive headroom.}
    \label{fig:proactivebench_composition}
\end{figure}

Figure~\ref{fig:proactivebench_composition} summarizes four properties of ProActEval: needs per scenario, facts per scenario, macro-domain coverage, and predictability structure.

%% file: main.bbl
\begin{thebibliography}{26}
\providecommand{\natexlab}[1]{#1}
\providecommand{\url}[1]{\texttt{#1}}
\expandafter\ifx\csname urlstyle\endcsname\relax
  \providecommand{\doi}[1]{doi: #1}\else
  \providecommand{\doi}{doi: \begingroup \urlstyle{rm}\Url}\fi

\bibitem[De~Min et~al.(2026)De~Min, Roy, Lathuili{\`e}re, Ricci, and
  Mancini]{de2026proactivebench}
Thomas De~Min, Subhankar Roy, St{\'e}phane Lathuili{\`e}re, Elisa Ricci, and
  Massimiliano Mancini.
\newblock Proactivebench: Benchmarking proactiveness in multimodal large
  language models.
\newblock \emph{arXiv preprint arXiv:2603.19466}, 2026.

\bibitem[Deng et~al.(2023)Deng, Lei, Lam, and Chua]{deng2023survey}
Yang Deng, Wenqiang Lei, Wai Lam, and Tat-Seng Chua.
\newblock A survey on proactive dialogue systems: Problems, methods, and
  prospects.
\newblock \emph{arXiv preprint arXiv:2305.02750}, 2023.

\bibitem[Drummond and Brough(2016)]{drummond2016proactive}
Suzie Drummond and Paula Brough.
\newblock Proactive coping and preventive coping: Evidence for two distinct
  constructs.
\newblock \emph{Personality and Individual Differences}, 92:\penalty0 123--127,
  2016.

\bibitem[Du et~al.(2024)Du, Wang, Zhao, Liang, Wang, Zhong, Wang, and
  Wong]{du2024perltqa}
Yiming Du, Hongru Wang, Zhengyi Zhao, Bin Liang, Baojun Wang, Wanjun Zhong,
  Zezhong Wang, and Kam-Fai Wong.
\newblock Perltqa: A personal long-term memory dataset for memory
  classification, retrieval, and fusion in question answering.
\newblock In \emph{Proceedings of the 10th SIGHAN Workshop on Chinese Language
  Processing (SIGHAN-10)}, pages 152--164, 2024.

\bibitem[Gao et~al.(2025)Gao, Geng, Hua, Hu, Juan, Liu, Liu, Qiu, Qi, Wu,
  et~al.]{gao2025survey}
Huan-ang Gao, Jiayi Geng, Wenyue Hua, Mengkang Hu, Xinzhe Juan, Hongzhang Liu,
  Shilong Liu, Jiahao Qiu, Xuan Qi, Yiran Wu, et~al.
\newblock A survey of self-evolving agents: On path to artificial super
  intelligence.
\newblock \emph{arXiv preprint arXiv:2507.21046}, 1, 2025.

\bibitem[Greenglass(1999)]{greenglass1999proactive}
Esther Greenglass.
\newblock The proactive coping inventory (pci): A multidimensional research
  instrument.
\newblock In \emph{International Conference of}, 1999.

\bibitem[Gupta et~al.(2024)Gupta, Kirtania, Singha, Gulwani, Radhakrishna,
  Soares, and Shi]{gupta2024metareflection}
Priyanshu Gupta, Shashank Kirtania, Ananya Singha, Sumit Gulwani, Arjun
  Radhakrishna, Gustavo Soares, and Sherry Shi.
\newblock Metareflection: Learning instructions for language agents using past
  reflections.
\newblock In \emph{Proceedings of the 2024 Conference on Empirical Methods in
  Natural Language Processing}, pages 8369--8385, 2024.

\bibitem[Hu et~al.(2024)Hu, Guo, Tang, Ma, Yao, Yang, and Xu]{hu2024designing}
Jiaxiong Hu, Jingya Guo, Ningjing Tang, Xiaojuan Ma, Yuan Yao, Changyuan Yang,
  and Yingqing Xu.
\newblock Designing the conversational agent: asking follow-up questions for
  information elicitation.
\newblock \emph{Proceedings of the ACM on Human-Computer Interaction},
  8\penalty0 (CSCW1):\penalty0 1--30, 2024.

\bibitem[Kim et~al.(2024)Kim, Chay, Hwang, Kyung, Chung, Cho, Jo, and
  Choi]{kim2024dialsim}
Jiho Kim, Woosog Chay, Hyeonji Hwang, Daeun Kyung, Hyunseung Chung, Eunbyeol
  Cho, Yohan Jo, and Edward Choi.
\newblock Dialsim: A real-time simulator for evaluating long-term dialogue
  understanding of conversational agents.
\newblock \emph{arXiv preprint arXiv:2406.13144}, 2024.

\bibitem[Liao et~al.(2023)Liao, Yang, and Shah]{liao2023proactive}
Lizi Liao, Grace~Hui Yang, and Chirag Shah.
\newblock Proactive conversational agents in the post-chatgpt world.
\newblock In \emph{Proceedings of the 46th international ACM SIGIR conference
  on research and development in information retrieval}, pages 3452--3455,
  2023.

\bibitem[Lin et~al.(2025)Lin, Snell, Wang, Packer, Wooders, Stoica, and
  Gonzalez]{lin2025sleep}
Kevin Lin, Charlie Snell, Yu~Wang, Charles Packer, Sarah Wooders, Ion Stoica,
  and Joseph~E Gonzalez.
\newblock Sleep-time compute: Beyond inference scaling at test-time.
\newblock \emph{arXiv preprint arXiv:2504.13171}, 2025.

\bibitem[Liu et~al.(2024)Liu, Huang, Zeng, Hao, Yu, Li, Wang, Gan, Liu, Yu,
  et~al.]{liu2024toolace}
Weiwen Liu, Xu~Huang, Xingshan Zeng, Xinlong Hao, Shuai Yu, Dexun Li, Shuai
  Wang, Weinan Gan, Zhengying Liu, Yuanqing Yu, et~al.
\newblock Toolace: Winning the points of llm function calling.
\newblock \emph{arXiv preprint arXiv:2409.00920}, 2024.

\bibitem[Lu et~al.(2024)Lu, Yang, Qian, Chen, Luo, Wu, Wang, Cong, Zhang, Lin,
  et~al.]{lu2024proactive}
Yaxi Lu, Shenzhi Yang, Cheng Qian, Guirong Chen, Qinyu Luo, Yesai Wu, Huadong
  Wang, Xin Cong, Zhong Zhang, Yankai Lin, et~al.
\newblock Proactive agent: Shifting llm agents from reactive responses to
  active assistance.
\newblock \emph{arXiv preprint arXiv:2410.12361}, 2024.

\bibitem[Packer et~al.(2023)Packer, Fang, Patil, Lin, Wooders, and
  Gonzalez]{packer2023memgpt}
Charles Packer, Vivian Fang, Shishir\_G Patil, Kevin Lin, Sarah Wooders, and
  Joseph\_E Gonzalez.
\newblock Memgpt: towards llms as operating systems.
\newblock 2023.

\bibitem[Park et~al.(2023)Park, O'Brien, Cai, Morris, Liang, and
  Bernstein]{park2023generative}
Joon~Sung Park, Joseph O'Brien, Carrie~Jun Cai, Meredith~Ringel Morris, Percy
  Liang, and Michael~S Bernstein.
\newblock Generative agents: Interactive simulacra of human behavior.
\newblock In \emph{Proceedings of the 36th annual acm symposium on user
  interface software and technology}, pages 1--22, 2023.

\bibitem[Shinn et~al.(2023)Shinn, Cassano, Gopinath, Narasimhan, and
  Yao]{shinn2023reflexion}
Noah Shinn, Federico Cassano, Ashwin Gopinath, Karthik Narasimhan, and Shunyu
  Yao.
\newblock Reflexion: Language agents with verbal reinforcement learning.
\newblock \emph{Advances in neural information processing systems},
  36:\penalty0 8634--8652, 2023.

\bibitem[Tan et~al.(2025)Tan, Zhang, Ma, Chen, Dai, and Dong]{tan2025membench}
Haoran Tan, Zeyu Zhang, Chen Ma, Xu~Chen, Quanyu Dai, and Zhenhua Dong.
\newblock Membench: Towards more comprehensive evaluation on the memory of
  llm-based agents.
\newblock In \emph{Findings of the Association for Computational Linguistics:
  ACL 2025}, pages 19336--19352, 2025.

\bibitem[Wang et~al.(2023{\natexlab{a}})Wang, Liang, Yang, Huang, Wu, Wu, Lu,
  Ma, and Li]{wang2023enhancing}
Bing Wang, Xinnian Liang, Jian Yang, Hui Huang, Shuangzhi Wu, Peihao Wu, Lu~Lu,
  Zejun Ma, and Zhoujun Li.
\newblock Enhancing large language model with self-controlled memory framework.
\newblock \emph{arXiv preprint arXiv:2304.13343}, 2023{\natexlab{a}}.

\bibitem[Wang et~al.(2023{\natexlab{b}})Wang, Xie, Jiang, Mandlekar, Xiao, Zhu,
  Fan, and Anandkumar]{wang2023voyager}
Guanzhi Wang, Yuqi Xie, Yunfan Jiang, Ajay Mandlekar, Chaowei Xiao, Yuke Zhu,
  Linxi Fan, and Anima Anandkumar.
\newblock Voyager: An open-ended embodied agent with large language models.
\newblock \emph{arXiv preprint arXiv:2305.16291}, 2023{\natexlab{b}}.

\bibitem[Wang et~al.(2025)Wang, Qian, Li, Qiu, Xue, Wang, Ji, and
  Wong]{wang2025toward}
Hongru Wang, Cheng Qian, Manling Li, Jiahao Qiu, Boyang Xue, Mengdi Wang, Heng
  Ji, and Kam-Fai Wong.
\newblock Toward a theory of agents as tool-use decision-makers.
\newblock \emph{arXiv preprint arXiv:2506.00886}, 2025.

\bibitem[Wang et~al.(2024)Wang, Ma, Feng, Zhang, Yang, Zhang, Chen, Tang, Chen,
  Lin, et~al.]{wang2024survey}
Lei Wang, Chen Ma, Xueyang Feng, Zeyu Zhang, Hao Yang, Jingsen Zhang, Zhiyuan
  Chen, Jiakai Tang, Xu~Chen, Yankai Lin, et~al.
\newblock A survey on large language model based autonomous agents.
\newblock \emph{Frontiers of Computer Science}, 18\penalty0 (6):\penalty0
  186345, 2024.

\bibitem[Wu et~al.(2024)Wu, Wang, Yu, Zhang, Chang, and Yu]{wu2024longmemeval}
Di~Wu, Hongwei Wang, Wenhao Yu, Yuwei Zhang, Kai-Wei Chang, and Dong Yu.
\newblock Longmemeval: Benchmarking chat assistants on long-term interactive
  memory.
\newblock \emph{arXiv preprint arXiv:2410.10813}, 2024.

\bibitem[Yan et~al.(2025)Yan, Li, Qian, Lu, and Liu]{yan2025general}
BY~Yan, Chaofan Li, Hongjin Qian, Shuqi Lu, and Zheng Liu.
\newblock General agentic memory via deep research.
\newblock \emph{arXiv preprint arXiv:2511.18423}, 2025.

\bibitem[Zhang et~al.(2026)Zhang, Zhang, Chen, Huang, Zheng, Wang, Guo, Mo,
  Bae, Zou, et~al.]{zhang2026lightweight}
Jiaquan Zhang, Chaoning Zhang, Shuxu Chen, Zhenzhen Huang, Pengcheng Zheng,
  Zhicheng Wang, Ping Guo, Fan Mo, Sung-Ho Bae, Jie Zou, et~al.
\newblock Lightweight llm agent memory with small language models.
\newblock \emph{arXiv preprint arXiv:2604.07798}, 2026.

\bibitem[Zhang et~al.(2024)Zhang, Dai, Chen, Jiang, Li, Zhu, Chen, Xie, Dong,
  and Wen]{zhang2024memsim}
Zeyu Zhang, Quanyu Dai, Luyu Chen, Zeren Jiang, Rui Li, Jieming Zhu, Xu~Chen,
  Yi~Xie, Zhenhua Dong, and Ji-Rong Wen.
\newblock Memsim: A bayesian simulator for evaluating memory of llm-based
  personal assistants.
\newblock \emph{arXiv preprint arXiv:2409.20163}, 2024.

\bibitem[Zhong et~al.(2024)Zhong, Guo, Gao, Ye, and Wang]{zhong2024memorybank}
Wanjun Zhong, Lianghong Guo, Qiqi Gao, He~Ye, and Yanlin Wang.
\newblock Memorybank: Enhancing large language models with long-term memory.
\newblock In \emph{Proceedings of the AAAI conference on artificial
  intelligence}, volume~38, pages 19724--19731, 2024.

\end{thebibliography}
